# ProBoost: a Boosting Method for Probabilistic Classifiers

Fábio Mendonça, Sheikh Shanawaz Mostafa, Fernando Morgado-Dias, Antonio G. Ravelo-García, and Mário A. T. Figueiredo

**Abstract**—ProBoost, a new boosting algorithm for probabilistic classifiers, is proposed in this work. This algorithm uses the epistemic uncertainty of each training sample to determine the most challenging/uncertain ones; the relevance of these samples is then increased for the next weak learner, producing a sequence that progressively focuses on the samples found to have the highest uncertainty. In the end, the weak learners' outputs are combined into a weighted ensemble of classifiers. Three methods are proposed to manipulate the training set: undersampling, oversampling, and weighting the training samples according to the uncertainty estimated by the weak learners. Furthermore, two approaches are studied regarding the ensemble combination. The weak learner herein considered is a standard convolutional neural network, and the probabilistic models underlying the uncertainty estimation use either variational inference or Monte Carlo dropout. The experimental evaluation carried out on MNIST benchmark datasets shows that ProBoost yields a significant performance improvement. The results are further highlighted by assessing the relative achievable improvement, a metric proposed in this work, which shows that a model with only four weak learners leads to an improvement exceeding 12% in this metric (for either accuracy, sensitivity, or specificity), in comparison to the model learned without ProBoost.

━━━━━━━━━━━━━━━━ ◆ ━━━━━━━━━━━━━━━━

## 1 INTRODUCTION

THE combination of multiple simple classifiers to produce a model whose performance surpasses a simple classifier is known as ensemble learning [1], [2]. This combination is often performed by methods that construct a weighted combination of the individual classifiers. The rationale behind these models is that training a combination of multiple simpler classifiers is easier and more effective than training a single complex classifier [2]. Boosting is a metaheuristic that exploits the strategy of building an ensemble of (so-called weak) classifiers, where each weak classifier (or weak learner) contributes to producing a more robust model [1]. There are two main design choices that boosting methods need to address: (i) how to adjust the training dataset for each weak learner; (ii) how to combine the weak learners to obtain a final stronger classifier [3].

The essence of boosting is to repeatedly use a base weak learner on differently weighted versions of the training data [4]. Therefore, selecting the proper weak learner is of the utmost importance when the goal is to improve the overall performance of the proposed model. In this regard, deep neural networks can provide excellent results for broad types of problems when compared to conventional machine learning algorithms, since they can learn data representations directly from data [5]. Another relevant factor is their scalability when trained by stochastic gradient descent or related methods, which can handle millions, even billions, of samples [6].

Among deep learning models, *convolutional neural networks* (CNN) have attained significant performance in numerous fields, especially on image data, as they can automatically and adaptively learn hierarchies of spatial features [7], [8]. However, CNNs usually require large amounts of training data and are prone to overfitting if the number of training samples is small, yielding classifiers that produce over-confident decisions [9]. Nevertheless, even if enough data is available, the uncertainty in a classifier's predictions is still a key element to be evaluated. Such analysis can be carried out by using Bayesian learning [10][11][12][13].

The limitation of standard deterministic deep learning, regarding uncertainty assessment, resides in the fact that a single point-estimate of the model weights is obtained. In contrast, Bayesian learning of neural networks produces estimates of the weights' distribution, thus allowing the predictive uncertainty to be divided into two components. One uncertainty component arises from the noise inherent in the data, while the other is related to the model uncertainty. These uncertainties are known as *aleatoric* and *epistemic*, respectively [14].

Epistemic uncertainty, also known as model uncertainty, is of particular interest as it can reflect the degree of confidence an estimated model has in its predictions. As a result, if a classifier displays a large uncertainty regarding which class a sample belongs to, then its decision should not be

- F.M. is with the Interactive Technologies Institute (ITI/LARSyS and ARDITI), 9020-105 Funchal, Portugal, and with the University of Madeira, 9000-082 Funchal, Portugal. E-mail: fabioruben@staff.uma.pt.
- S.S.M. is with the Interactive Technologies Institute (ITI/LARSyS and ARDITI), 9020-105 Funchal, Portugal. E-mail: sheikh.mostafa@tecnico.ulisboa.pt.
- F.M-D. is with the Interactive Technologies Institute (ITI/LARSyS and ARDITI), 9020-105 Funchal, Portugal, and with the University of Madeira, 9000-082 Funchal, Portugal. E-mail: morgado@uma.pt.
- A.G.R-G. is with the Interactive Technologies Institute (ITI/LARSyS and ARDITI), 9020-105 Funchal, Portugal, and with the Institute for Technological Development and Innovation in Communications, Universidad de Las Palmas de Gran Canaria, 35001 Las Palmas de Gran Canaria, Spain. E-mail: antonio.ravelo@ulpgc.es.
- M.A.T.F. is with the Instituto de Telecomunicações, Instituto Superior Técnico, Universidade de Lisboa, 1049-001 Lisboa, Portugal. E-mail: mario.figueiredo@tecnico.ulisboa.pt.

2trusted. The samples about which a classifier has the highest uncertainty are likely the most challenging to classify. Inspired by this observation and recalling the boosting idea of combining multiple weak learners, a novel methodology is proposed in this work, with the following rationale: identify the hardest (most uncertain) samples in the training set and increase their relevance in training subsequent classifiers in the ensemble, thus producing a sequence of weak classifiers that are increasingly specialized on the samples found to be the most uncertain by the previous ones. This scheme leads to the creation of several weak learners focused on a subset of the data according to uncertainty. The hypothesis proposed in this work is that combining an ensemble of classifiers (weak learners) trained on different subsets of the training dataset that are selected by a sequence of uncertainty assessments yields a better model than any individual classifier in the ensemble.

This paper proposes *ProBoost*, a boosting method that exploits the ability to obtain uncertainty estimates for individual samples using Bayesian learning. The proposed algorithm is instantiated with CNNs as the weak learners and experimentally assessed on standard datasets contaminated with noise. Two different approaches for producing the probabilistic classifier are evaluated to determine which is more suitable *ProBoost*: *variational inference* (VI) [15] and *Monte Carlo* (MC) *dropout* (MCD) [16]. For each approach, three methods were tested to control the relevance of the training samples for the subsequent weak learners: (i) keeping only the hardest samples); (ii) duplicating the hardest samples; (iii) changing the weight of the hardest samples.

The rest of this article is organized as follows: an overview of boosting methods is presented in Section 2. The building blocks underlying the proposed method are presented in Section 3. The proposed *ProBoost* algorithm is presented in Section 4. The experimental results are presented and discussed in Section 5. Finally, Section 6 concludes the article.

## 2 RELATED WORK

Several boosting algorithms have been proposed, differing essentially on the methodology used to produce the differently weighted versions of the training data [2]. The first boosting algorithm, presented by Schapire [17], randomly splits the training set into three partitions and performs an analysis based on three classifiers (using a majority voting scheme to combine the classifiers' outputs). Freund and Schapire [18] proposed the well-known AdaBoost (*adaptive boosting*) algorithm, whose rationale is to use multiple weighted versions of the same training set. The algorithm trains a sequence of weak learners, with the training weights depending on the accuracy of the previous classifiers in the sequence.

Several variants of AdaBoost have been proposed for both binary and multiclass problems. For binary classification, one of the first proposals was RealBoost, which uses the classifier outputs (before thresholding) [19]. A similar approach, named Gentle AdaBoost [20], uses weighted class probabilities. Other variants, such as LogitBoost [20], aim to minimize a specific loss, such as the logistic loss (negative conditional log-likelihood). Other algorithms, like AnyBoost [21] and Asymmetric Boosting [22], for example, use gradient descent to choose a linear combination of elements.

There are variants of AdaBoost developed for specific purposes, such as Modest AdaBoost [23], which tries to attain a lower generalization error by using a different weighting scheme for the incorrectly and the correctly classified data. Other examples include ActiveBoost [24], which uses active learning to lessen the undesirable effect of noisy data, and MadaBoost [2], which also mitigates problems related to noisy data. Methods, such as Jensen-Shannon Boosting [25], were developed to improve the dissimilarity between two classes, by using the Jensen-Shannon divergence. On the other hand, algorithms such as SoftBoost [26] and WeightBoost [27] try to diminish AdaBoost's overfitting problem. In contrast, methods like ERLPBoost [28] impose an iteration bound while maximizing the margin between samples from different classes.

The AdaBoost.M1 and AdaBoost.M2 variants [29] have been proposed as extensions of the standard AdaBoost for multiclass problems. The first variant weights the weak learners according to the error rate, while the second variant increases the sample's weights by examining a loss. Several authors proposed variations of these algorithms, such as AdaBoost.M1W [30], which uses a different approach than AdaBoost.M1 for weighting the weak learners, and BoostMA [31], which modified AdaBoost.M2 to minimize a different error. AdaBoost.MR [19] is also a variant based on ranking loss. Other algorithms, *e.g.*, AdaBoost.SEEC [32], use stage-wise gradient descent with a regularization parameter. Finally, Multi-class AdaBoost [33] can lower the misclassification error rate by minimizing a specific loss function.

A considerably dissimilar approach is followed in AdaBoost.BHC [34], which uses $\beta - 1$ classifiers for a problem with $\beta$ classes. On the other hand, AdaBoost.HM [35] combines multiple multiclass classifiers using a proposed hypothesis margin. Boosted trained jointly *one-versus-all* (OvA) classifiers is the approach followed by JointBoost [36], and it was shown that this method can outperform independently trained OvA classifiers.

Despite the extensive literature and large amount of work, to the best of the authors' knowledge, no boosting method has been proposed for probabilistic models, exploiting the availability of sample-wise epistemic uncertainty estimates. Such a gap is addressed in this work, by proposing *ProBoost*, a new algorithm that encompasses the core rationale of boosting while using a novel view on how to select the most challenging samples to form the training dataset for the subsequent weak learners.

## 3 BUILDING BLOCKS

This section reviews the building blocks underlying our proposed approach, using convolutional neural networks (CNNs) as weak learners, and the two methods used to estimate uncertainty: variational inference (VI) and Monte Carlo dropout (MCD).

## 3.1 CNNs as Weak Learners

In this work we focus on (two-dimensional) image classification tasks, thus CNNs, known for their adequacy to image classification tasks [37], are a natural choice for the weak learners. Moreover, since the purpose is not to attain the highest possible performance with a single classifier, a standard CNN architecture was selected. More specifically, a variation of the well-known LeNet-5 [38] model (described in detail in the next paragraph) was used as the weak learner.

The following variation of LeNet-5 was used as weak learner. The first layer is convolutional, using six kernels, each with size five, stride one, zero padding, and ReLU as activation function. The second layer performs max-pooling, using pooling size and stride of two in both directions, and zero padding. The third layer is similar to the first one, but with 16 kernels. The fourth layer is equal to the second one. The fifth layer is similar to the first and third layers, but with 120 kernels. The sixth layer is fully connected (dense), receiving a flattened input, and has 84 neurons with ReLU activation function. The seventh (and final) layer is fully connected and performs the classification, with ten output classes with softmax activation.

## 3.2 Variational Inference

In a simplistic view, using a Bayesian approach would only require following the Bayes theorem, sampling the posterior, and then performing model averaging. However, except in very simple scenarios, sampling from the posterior is known to be extremely difficult, or even unfeasible [12][13]. For the specific case of neural networks, the posterior distribution $P(W|D)$ of the weights $W$, given the training data $D$, could be used to obtain a Bayesian forecast of the label $y$ for new data point $x$ by taking the posterior expectation with respect to $W$ [12][13][39],

$$P(y|x,D) = E_{P(W|D)}[P(y|x,W,D) \mid D]$$
$$= \int P(y|x,W) \, P(W|D) \, dW \,, \quad (1)$$

where $P(y|x,W)$ is the predicted probability that x belongs to class $y$, given by a network with parameters $W$. Therefore, all possible values for the weights $W$, weighted by the posterior distribution, contribute to predicting the class of observation $x$. This view is equivalent to using an infinite ensemble of neural networks, and the expectation in (1) can only be analytically computed in extremely simple cases, making this approach unfeasible in essentially all realistic applications [39]. One possible alternative would be to approximate (1) by a sample average,

$$\hat{P}(y|x,D) \approx \frac{1}{n} \sum_{i=1}^{n} P(y|x,W^i) \,, \quad (2)$$

where $W^1, \ldots, W^n$ are independent and identically distributed (i.i.d.) samples from the parameter posterior $P(W|D)$. However, even in the case where the posterior can be computed, it is still problematic to sample from it directly in non-trivial models, due to the (very) high dimensionality of the parameter space.

Multiple algorithms have been proposed as alternatives to sampling directly from the posterior. The most frequently used is *Markov chain Monte Carlo* (MCMC), which approximates the sampling process from the posterior and VI, which approximates the posterior with a tractable form.

MCMC generates a sequence of random samples by constructing a Markov chain whose stationary distribution should match the posterior distribution. However, a long time is usually needed to attain proper convergence and, usually, several aspects need to be carefully tuned, making MCMC a method that is hard to use in practice with deep neural networks [40].

Variational inference (VI) is used in statistics and machine learning to cast the problem of finding a probability distribution as an optimization problem. The rationale behind VI is to select a member $q$ from a family $Q$ of distributions parametrized by $\theta$, i.e., $Q = \{q(W|\theta), \theta \in \Theta\}$ (where $\Theta$ is the set of valid parameters), that is (desirably) close to the target distribution (in our case, the posterior $P(W|D)$), using the *Kullback-Leibler divergence* (KLD) to measure the distance [15]. VI is formulated as the optimization problem of finding the parameters $\vartheta$ that make $q(W|\theta)$ as close as possible (in KLD sense) to the conditional of interest $P(W|D)$ [15],

$$q^*(W|\theta) = arg \min_{q(W|\theta) \in Q} KLD[q(W|\theta)||P(W|D)] \,; \quad (3)$$

equivalently, $q^*(W|\theta) = q(W|\theta^*)$, where

$$\theta^* = arg \min_{\theta \in \Theta} KLD[q(W|\theta)||P(W|D)] \,. \quad (4)$$

As $P(W|D)$ is unknown, it is not possible to directly solve this problem, and it is necessary to optimize an approximation thereof. The standard approach is to use the so-called *evidence lower bound* (ELBO) [39], the maximization of which implies the minimization of an upper bound on the KLD in (3):

$$ELBO(\theta) = E_{q(W|\theta)}[log[P(D|W)]] \\ - KLD[q(W|\theta)||P(W)]. \quad (5)$$

The second term in (5) is prior-dependent; it is typically seen as the complexity cost, and it encourages solutions whose densities are close to the prior. The first term is data-dependent; it is usually named the expected log-likelihood, as it corresponds to the (log of the) probability of the data given the model parameters $W$, encouraging solutions that explain well the observed data. As a result, the ELBO embodies a trade-off between satisfying the prior $P(W)$, which usually expresses a preference for some sort of simplicity, and fitting the data, which usually calls for complex models [39]. Although Bayesian approximation can be performed by VI, using a prior distribution for the model weights to produce a regularization effect, the exact computation of the ELBO as presented above is computationally prohibitive. A common approach is to use a Gaussian homoscedastic approximation, where each element of $Q$ is a product of identical normal densities, thus with only two parameters: mean and variance [41].

A sample-based approximation to the ELBO can be used to optimize the neural network via gradient descent on this loss function. This approximation is given by [42]

$$ELBO(\theta) \approx \sum_{i=1}^{n} \log[q(W^i|\theta)] - \log[P(W^i)] - \log[P(D|W^i)], \quad (6)$$

where $n$ is the number of MC samples drawn from $q$.

A common approach for training neural networks with VI is based on weight perturbation techniques, which stochastically sample the weights of the network at training time. However, weight perturbation usually produces high-variance gradient estimates, since all mini-batch samples share the same perturbation (restricting the variance reduction effect produced when using large mini-batches). The *flipout* estimator [43] was introduced to lessen this problem, decorrelating the gradients within each mini-batch, by sampling pseudo-independent weight perturbations for each sample. Specifically, a base perturbation $\widehat{W}$ is shared by all samples $s$ in each mini-batch, but each sample is perturbed by the result of element-wise multiplication (denoted ∘) of $\widehat{\Delta W}$ (stochastic perturbation) with a rank-one sign matrix as [43]

$$\Delta W_s = \widehat{\Delta W} \circ a_s b_s^T, \quad (7)$$

where $a$ and $b$ are random vectors sampled uniformly from $\{-1,+1\}$; as a result, *flipout* yields an unbiased estimator of the loss gradients. By using the flipout estimator, the output of the forward pass on a layer, for all examples in the mini-batch ($X$), is given by [43]

$$Y = \Phi[X\overline{W} + [(X \circ B)\widehat{\Delta W}] \circ A], \quad (8)$$

where $\overline{W}$ are the average weights. Since $A$ and $B$ were independently sampled, it is possible to backpropagate with respect to $X$, $\widehat{\Delta W}$, and $\overline{W}$.

### 3.3 Monte Carlo Dropout

The conventional use of dropout is to operate as a regularization procedure during training, by applying multiplicative noise (typically Bernoulli, *i.e.*, taking values in {0, 1}) to the previous layer [44]. However, if dropout is used during test time, some features from the previous layer will be set to zero, leading to a network output that is not affected by some of the features. Therefore, every time the test data is examined, different outputs will be produced. Hence, by using random dropout, the network outputs can be viewed as samples from a probability distribution.

Gal and Ghahramani [45] showed that MCD can provide an approximation to a Bayesian posterior, if a considerably high dropout rate is used with a sufficient number of predictions. This approach usually requires the use of dropout for each layer with tunable weights (except for the last layer). Thus, MCD samples binary variables for each network unit except in the output layer. The result is a weight distribution, which can be either 0 or the weight value. The estimated posterior probability of class $y$ for some test sample $x_{test}$ is thus obtained in the same way as for VI, where the prediction is computed by averaging over $T$ sampled assessments through [41],

$$P(y|x_{test}, D) = \frac{1}{T}\sum_{t=1}^{T} P(y|x_{test}, W^t), \quad (9)$$

where $W_t$ is the weight configuration in the t-th dropout sample.

## 4 PROPOSED METHOD: *PROBOOST*

### 4.1 Rationale and Outline

The idea of *ProBoost* is to sequentially train weak learners that are progressively specialized on the most difficult data points, that is, those exhibiting the highest epistemic uncertainty. We propose using a cascade/sequential training procedure, where each new weak learner can be seen as a new level. Apart from the first, each level receives information from the previous level regarding which data is more challenging (has the highest uncertainty). As a result, each new level has the combined information from all previous levels regarding which data is the most difficult, focusing each new weak learner on the hardest data.

Bayesian learning methods are capable of expressing and assessing epistemic uncertainties since they compute (or approximate) a posterior density on the model parameters. For example, assuming that

$$\frac{1}{T}\sum_{t=1}^{T} P(y|x_{test}, W^t) \approx P(y|x_{test}, D), \quad (10)$$

the total predictive variance for some test sample $x_{test}$ is given by

$$\widehat{Var}[P(y|x_{test})] = \\ = \sum_{y} \frac{1}{T}\sum_{t=1}^{T}[P(y|x_{test}, W^t) - P(y|x_{test}, D)]^2. \quad (11)$$

Since the epistemic uncertainty reflects the degree of confidence in the model predictions, a sample with higher epistemic uncertainty can be viewed as a harder sample for the classifier than a sample with lower epistemic uncertainty. The core idea underlying *ProBoost* is to stress these harder samples for the subsequent levels of the cascade, leading to a sequential information flow concerning which samples are the hardest, from the first to the last level. As a consequence, if many levels are used, the last levels will likely focus on the most difficult samples. In fact, these hardest samples may be outliers, suggesting that *ProBoost* could conceivably be used for outlier detection, a research direction that we leave for future work.

Three methods were studied to implement *ProBoost*, leading to three different approaches regarding the training dataset.

1) The first method, called *undersampled ProBoost,* evaluates the epistemic uncertainty at each level, discarding (a fraction $R$ of) the samples with the lowest uncertainty before passing the dataset to the next level, such that the amount of data in the last level is 25% ($\tau = 0.25$) of the initial training set size. This percentage was empirically chosen as it was observed that when the initial training dataset is small, using a too low value can produce a dataset with insufficient samples to train the weak learner of the last level. Of course, this fraction can be adjusted for other applications or other datasets. The reduction factor $R$ that leads to the desired amount of data (length of the dataset) in the last level can be easily computed: for a system with $V$ levels (thus $V-1$ reduction steps) and reduction factor $R$, we have $\tau = R^{V-1}$; solving for $R$ leads to $R = \sqrt[V-1]{\tau}$. For example, for a system with 4 levels to have 25% of the data in the fourth level, we need $R = \sqrt[3]{0.25} \approx 0.63$.

2) The second method, referred to as *oversampled ProBoost,* uses a growth factor $G$ that controls how many samples (with the highest uncertainty) are duplicated before passing the training dataset to the next level. This factor is selected so that the increase in the amount of data from each level to the next is 25%, i.e., $G = 1.25$, thus the length of the dataset at level $V$ is $G^{V-1}$. The reasoning behind this method is to increase the hardest samples prevalence in the next levels, by changing the data distribution, but allowing all levels to learn from all data. Clearly, this method leads to an exponential growth of the dataset size with the number of levels, making it usable only for low $V$ and/or small datasets.

3) The third method, termed *weighted ProBoost,* is similar to the second one, but instead of duplicating the hardest samples, only their weight (in the training loss) is increased by one, keeping the same training dataset size on all levels. The second and third methods are equivalent if the entire dataset is loaded at once (batch training), but when using stochastic or mini-batch training, the results will likely differ.

Since the third method keeps the dataset size, it is suitable for larger values of $V$, requiring a similar training time for each level. Conversely, it is possible that method 1) is more appropriate when the training dataset is larger as it requires a lower number of computations. In all three methods, after the new training set if produced, it is shuffled and passed to the next level. An example of the *ProBoost* method's procedure for a model with three levels is presented in Fig. 1 of the supplementary material.

The structure of the algorithm is shown in Fig. 1, with an example based on the classic Iris dataset. Two features (sepal and petal lengths) were chosen for display purposes. The weak learners have a single dense layer (with *flipout* estimator) with a 3-class softmax activation, and 3 levels were used with *weighted ProBoost*. In the figure, it is possible to observe an inversed relationship between the sample weights and their uncertainty from levels 1 to 3, indicated by the marker size (larger size/weight denotes more uncertainty). This is particularly noticeable when samples from classes 1 and 2 overlap, where the boundary samples are identified as the most uncertain. In contrast, the class-0 samples, which are easily separable from the other classes, never increased their weights, since they were never in the 25% most uncertain, as expected.

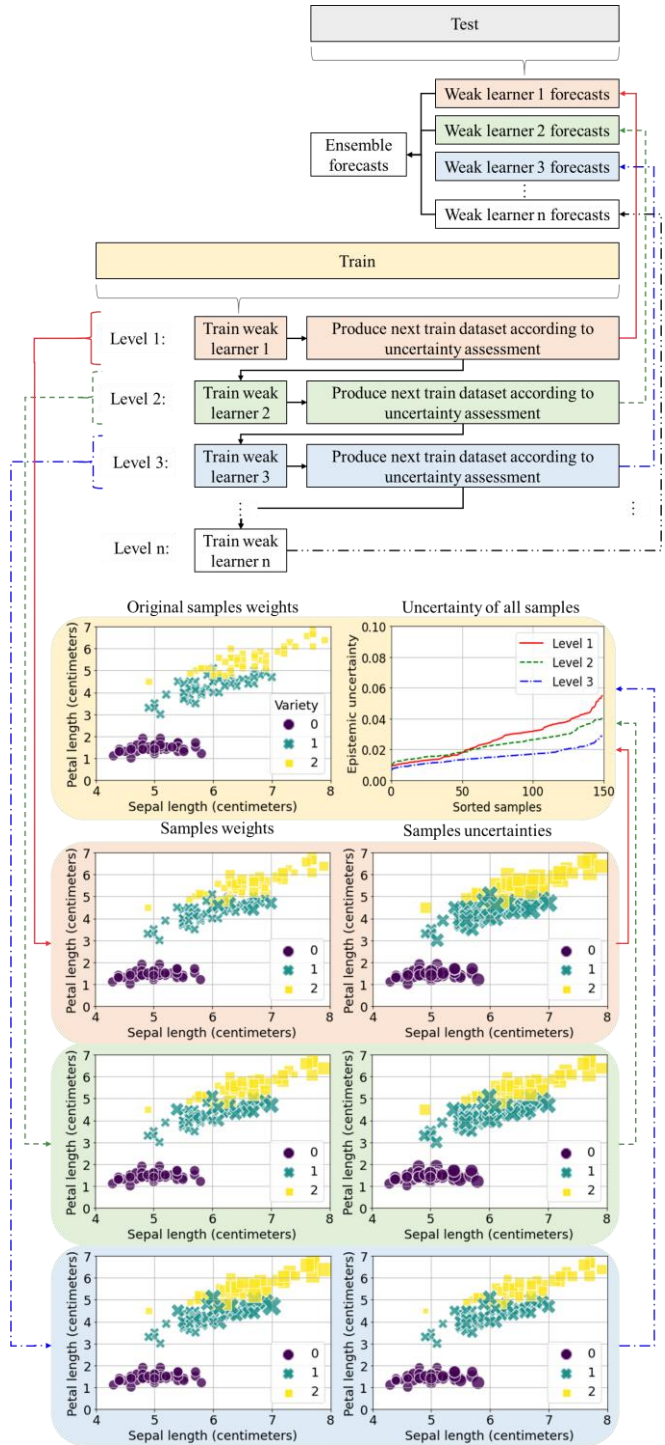

Fig. 1. Structure of the *ProBoost* algorithm, illustrated with application to the Iris dataset (see text for details).

## 4.2 Combining the Classifiers

The *ProBoost* algorithm produces a set of weak learners (each corresponding to a level), which were trained in different versions of the original training dataset. This implies that each weak learner produces different class probability estimates when presented with the same input. In the end, these estimates need to be combined using an ensemble approach. We chose to use a weighted sum rule to allow the combination to be performed directly on the classifier's softmax output (from zero to one), leading to the creation of the composite hypothesis. This ensemble method was selected as it has been shown to outperform other classifier combination schemes while being the most resilient to estimation errors of each weak learner [46].

The selected implementation for the ensemble considers a weighted sum of each weak learner's output. As a result, the output of the model is given by computing the arg max [6] to the ensemble output through

$$\hat{y}(x_{test}) = \arg\max_y \sum_{v=1}^{V} P(y|x_{test}, v)\, \psi_v, \tag{12}$$

where $\psi_v$ is the weight of the classifier/learner from level $v$. This can be viewed as a simplification of the Bayesian decision rule by relying on decision-support computations performed by the multiple weak learners.

Two approaches were examined regarding the selection of the weights $\psi_v$. The first uses constant values: unit weight for the first level ($\psi_1 = 1$), as it has seen the entire dataset, thus having the ability of producing broader hypotheses; all other levels have a weight of 0.5. The reasoning behind this decision was that these levels are trained on altered versions of the original dataset, thus can be regarded as less suitable for the common examples while being better for samples found to be the most uncertain. For the second approach, the weight of each weak learner is equal to its average accuracy over the whole training dataset. Therefore, the second approach allows the model to adapt the ensemble output to each weak learner.

### 4.2 Detailed Algorithms

The pseudo-code for the three variants of *ProBoost*, *undersampled*, *oversampled*, and *weighted*, are presented in Algorithm 1, 2, and 3, respectively.

The inputs to the *ProBoost* algorithms are the training dataset, the weak learning algorithm *wLearner* used as a base classifier for each level, an integer value $v$ to specify how many levels should be generated (the minimum number is two as choosing one leads to a single classifier model), the fraction parameter $\tau$, and the number of MC samples *MCS* used to compute the epistemic uncertainty. All instances of the training dataset initiate with the same unit weights. After the algorithm finishes, each of the obtained classifiers are combined as explained above.

### 4.3 Implementation Details

All models are implemented in Python 3 using the Tensorflow 2 platform and the Tensorflow Probability libraries. In the VI implementation, a Gaussian distribution with trainable parameters was employed for the posterior density of both the parameters of the convolutional layers and of the dense layers. The bias terms had a multivariate standard Gaussian distribution. These choices were made to keep the default parameters in the Tensorflow layers as this work aims to evaluate the *ProBoost* algorithm with standard weak learners.

**Algorithm 1.** Pseudo-code for undersampled *ProBoost* (method 1).

$d \leftarrow trainDataset$
$divValue \leftarrow 1 / (1 - \tau \wedge (1 / (v - 1)))$
**For** *level* = 1, 2, …, $v - 1$ **do**:
    $classifier_{level} \leftarrow$ train (*wLearner*, $d$)
    **For** $m$ = 1, 2, …, *MCS* **do**:
        $output_m \leftarrow$ test ($classifier_{level}$ ($d$))
    $u \leftarrow$ variance ($output_{1\ to\ MCS}$)
    $s \leftarrow$ sort $d$ according to ascending $u$
    $segment \leftarrow s$ [**from** integer (length ($s$) / $divValue$) **to** end]
    $d \leftarrow$ shuffle ($segment$)
$classifier_v \leftarrow$ train (*wLearner*, $d$)

**Algorithm 2.** Pseudo-code for oversampled *ProBoost* (method 2).

$d \leftarrow trainDataset$
**For** *level* = 1, 2, …, $v - 1$ **do**:
    $classifier_{level} \leftarrow$ train (*wLearner*, $d$)
    **For** $m$ = 1, 2, …, *MCS* **do**:
        $output_m \leftarrow$ test ($classifier_{level}$ ($d$))
    $u \leftarrow$ variance ($output_{1\ to\ MCS}$)
    $s \leftarrow$ sort $d$ according to ascending $u$
    $segment \leftarrow s$ [**from** integer (length ($s$) × (1 − $\tau$)) **to** end]
    $combined \leftarrow$ concatenate ($d$, $segment$)
    $d \leftarrow$ shuffle ($combined$)
$classifier_v \leftarrow$ train (*wLearner*, $d$)

**Algorithm 3.** Pseudo-code for weighted *ProBoost* (method 3).

$d \leftarrow trainDataset$
$sWeights =$ ones (length ($d$))
**For** *level* = 1, 2, …, $v - 1$ **do**:
    $classifier_{level} \leftarrow$ train (*wLearner*, $d$)
    **For** $m$ = 1, 2, …, *MCS* **do**:
        $output_m \leftarrow$ test ($classifier_{level}$ ($d$))
    $u \leftarrow$ variance ($output_{1\ to\ MCS}$)
    $d \leftarrow$ sort $d$ according to ascending $u$
    $sWeights \leftarrow$ sort $sWeights$ according to ascending $u$
    $sWeights$ (**from** integer (length ($d$) × (1 − $\tau$)) **to** end) $\leftarrow sWeights$ (**from** integer (length ($d$) × (1 − $\tau$)) **to** end) + 1
    $sWeights, d \leftarrow$ shuffle ($sWeights, d$)
$classifier_v \leftarrow$ train (*wLearner*, $d$)

The number of MC samples to compute the average of the VI models' results and check the epistemic uncertainty was selected to be 50 (independent runs of the estimator), a value indicated by Wen et al. [43]. For MCD, the dropout probability in every layer was 0.3, a value empirically based on examining state-of-the-art works, which recommend values ranging from 0.2 [16] to 0.5 [45], while the number of MC samples was increased to 200 to properly estimate the posterior. For both VI and MCD, ADAM was used as the optimization algorithm, with a batch size of 16 samples, and a maximum number of epochs of 300.

## 5 EXPERIMENTAL RESULTS AND DISCUSSION

### 5.1 Performance Metrics

Standard metrics were employed to experimentally evaluate the performance of the model predictions against the ground truth: *accuracy* (Acc), *sensitivity* (Sen), *specificity* (Spe), *positive predictive value* (PPV), and *negative predictive value* (NPV) [47]. Additionally, the diagnostic ability of the classifier was assessed by the *area under the receiver operating characteristic curve* (AUC) [48], using a OvA approach. The dataset provider previously established both training and testing subsets.

During training, an early stopping procedure was used to avoid overfitting, using a patience value of 10, and 30% of the training dataset as validation. The evaluation procedure was repeated ten times in each analysis to estimate the corresponding *mean* (μ) and *standard deviation* (σ). Furthermore, the *minimum* (∧), *maximum* (∨), and 95% *confidence interval* (CI) for the results were computed. The significance (p-value) of the results attained by the models using *ProBoost* against the single-classifier model (which does not use *ProBoost*) was assessed by the paired t-test (compare the means attained on the same test dataset under two separate scenarios). A one-tailed test was used as it is only relevant in one direction: an improvement from a single classifier to the models using *ProBoost* [49]. The results are considered significant if the p-value is lower than 0.05.

The notion of *relative obtainable improvement* (ROI) is proposed and used in this work. This metric allows the comparison of two scenarios, say A and B, making it possible to assess the absolute achievable improvement of scenario B against scenario A for a specific parameter in relative decrease of the misclassification error. The rationale for this metric is to evaluate how much improvement was attained by using the *ProBoost* compared to using a single classifier (in one case, to assess the improvement gained by using *ProBoost* with 10 levels when compared to using *ProBoost* with 4 levels). Hence, scenario A is without *ProBoost*, while scenario B is using *ProBoost*. The ROI is defined as

$$ROI = \frac{result\ B - result\ A}{1 - result\ A}.\quad(13)$$

As an example of the usefulness of ROI, consider two scenarios. In the first, a single classifier attains an Acc of 60%, which improves to 61% using *ProBoost*. For the second scenario, the single classifier and the model using *ProBoost* reach Acc of 98% and 99%, respectively. Arguably, the improvement in the second scenario is considerably more challenging to attain than the improvement in the first scenario, as the performance is closer to the maximum possible. However, in absolute terms, it is a 1% improvement in both scenarios. This difficulty in obtaining the improvement is highlighted in the ROI, as for the first and second scenarios, the ROI values are 2.5% and 50%, respectively. This indicates that the improvement in the second scenario is 50% of the possible increase to 100%, while in the first scenario, only 2.5% of the potential increase to 100% is attained. Therefore, it is defensible that the improvement in the second scenario is more relevant (and likely more difficult to obtain) than the first scenario's improvement.

### 5.2 Methodology

The experimental methodology followed is presented in Fig. 2. Two datasets were prepared from the standard MNIST and fashion-MNIST datasets. One of the produced datasets was used to assess the effect of having strong noise in the data, while the other was employed to observe the effect of having only part of the samples from the dataset contaminated with noise.

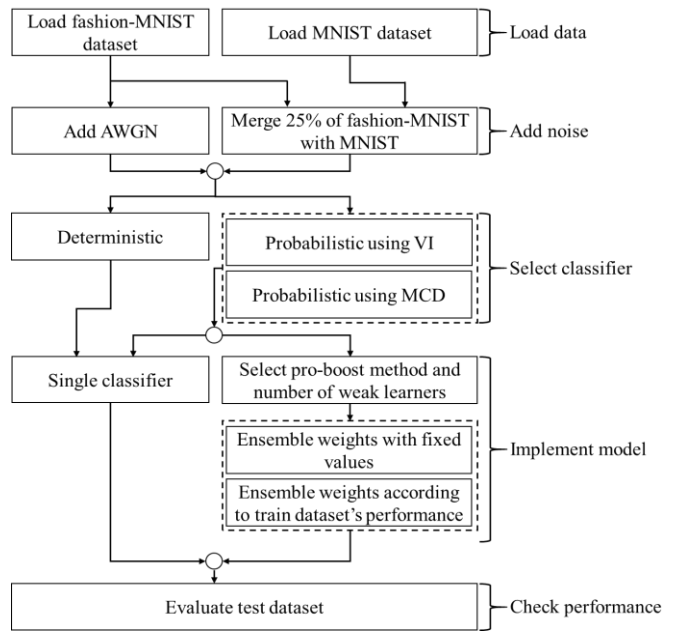

Fig. 2. Block diagram of the methodology employed in this work.

Three tests were carried out. The first test considers the first prepared dataset (fashion-MNIST dataset contaminated with additive white Gaussian noise - AWGN), evaluating the performance of *ProBoost* with one, two, three, and four levels. A deterministic model and the probabilistic models with a single classifier were also used as baselines. This first test was intended to investigate the potential of *ProBoost*, checking if there is an improvement when more levels are added.

The second test considers only models using *ProBoost* with four levels and a deterministic model. This test examines the second prepared dataset (fashion-MNIST with superimposed images from the MNIST digit dataset) and aims at assessing the performance of the best models using *ProBoost* with a dataset partially affected by noise.

The third test compares the weighting schemes employed for the ensemble, checking if they are too far from the best possible weights, taking the number of levels to 10. Hence, for this final test, weighted *ProBoost* was used, as this method was found to be the best in the second test. A comparative analysis was made between the results of the model using the weighting schemes previously described and the ideal (but not usable) weights selected by maximizing the accuracy on the test dataset.

The results presented in this section refer to the macro performance. The approaches examined for selecting the ensemble weights were denoted as FW for fixed weights and VW for variable weights.

## 5.3 Standard Datasets

Two datasets were used, both composed of 28x28 gray-scale images in ten different classes: fashion-MNIST (images of clothes), which has 60000 and 10000 samples for training and testing, respectively; MNIST (images of hand-written numbers), which has the same numbers of samples as fashion-MNIST. The images in both datasets are gray-scale, with pixel values in the [0, 255] range.

To test the effect of having strong noise in the data, making learning a classifier more difficult, the first dataset was prepared by contaminating the fashion-MNIST with AWGN, with mean and variance equal to 255/2. An example of a resulting sample is presented in Fig. 3.

The second prepared dataset contains all samples from fashion-MNIST, but in which 25% of the samples (randomly selected) were superimposed with images of the MNIST dataset (subsequently renormalizing the resulting images with the min-max algorithm [6]). This dataset was prepared to evaluate the effect of having only a small part of the samples contaminated with non-random noise, where each class is affected by a different pattern (numbers). The MNIST dataset samples were selected to be the same class as the fashion-MNIST dataset. For example, if a sample from the fashion-MNIST dataset belonging to class 6 (representing a shirt) was selected to be in the 25% group with noise, then a MNIST sample from class 6 (a hand-written 6) was selected; the result is an image of a hand-written six superimposed on a shirt, as illustrated in Fig. 3.

The attained Acc, PPV, and NPV for *ProBoost* with four levels are shown in Fig. 4 (a). The results show that *ProBoost* (all three examined methods, with two, three, or four levels) leads to significant improvements (p-value lower than 0.05) in all evaluated performance metrics. The detailed results are shown in Tables 1 and 2 of the supplementary material. The performance of the deterministic model is shown in Table 3 of the supplementary material.

Regarding the VI model, it is possible to conclude that the most substantial improvement is reached when changing the model from one level (no boosting) to two levels, keeping a similar performance when the number of levels is increased to three and four. On the other hand, for both *oversampled* and *weighted ProBoost*, performance consistently improves as the number of levels increases. These results suggest that when the easier samples are deleted (*undersampled ProBoost*), it is not worth going beyond two levels, arguably because the elimination process produces weak learners that are too focused on the hardest samples (successively more specialized as the number of levels increases), losing context on the general samples. For both *oversampled* and *weighted ProBoost*, the context is kept, and the importance of the hardest samples is increased. Hence, it is likely to produce weak learners that are progressively more focused on the hardest samples (as the number of levels increases) but still retain the context of the general samples. Another relevant point is the performance of *weighted ProBoost*, which was, on average, superior to oversampled *ProBoost*. These results are likely caused by the use of mini-batches, which favor the sample weight change (performed by *weighted ProBoost*) over the oversampling approach (used by *oversampled ProBoost*). These analyses are valid for both the FW and VW weighting schemes. However, FW seems to be a better option when using *undersampled ProBoost*, while VW was superior for both *oversampled and weighted ProBoost*.

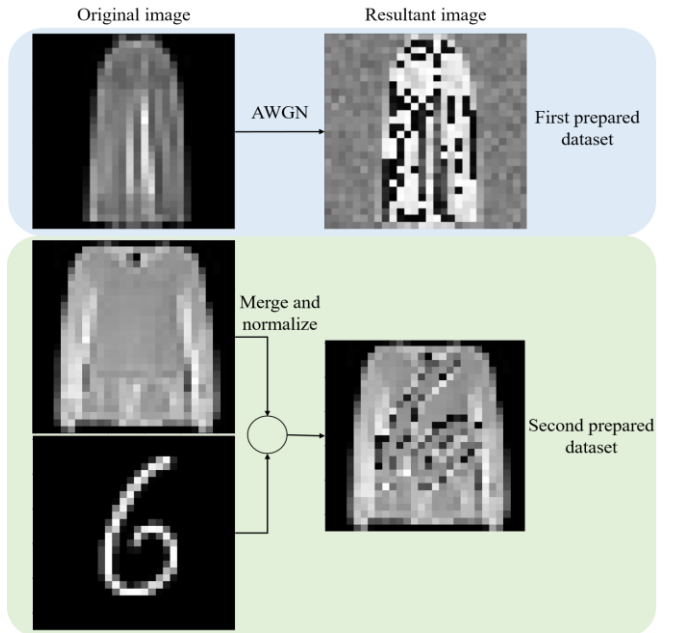

Fig. 3. Examples of a sample from the first and second prepared dataset depicting the original and resultant images.

It is possible to observe that VI-based *ProBoost* reaches a superior performance than MCD, which can mean that MCD leads to a less accurate epistemic uncertainty estimation, affecting the performance of *ProBoost*. Nevertheless, yields significant performance improvement for all tested MCD models (p-value < 0.05). Furthermore, *undersampled ProBoost* has its peak performance with two levels, with more levels leading to worst results. This is likely due to the same reasons identified for VI: increasing the number of levels leads to models too focused on only part of the samples. If the uncertainty assessment was less precise, the new levels might not be learning in the most challenging samples. For both *oversampled* and *weighted ProBoost*, the results are similar to the VI-based models, and the same interpretation applies, as well as for FW against the VW.

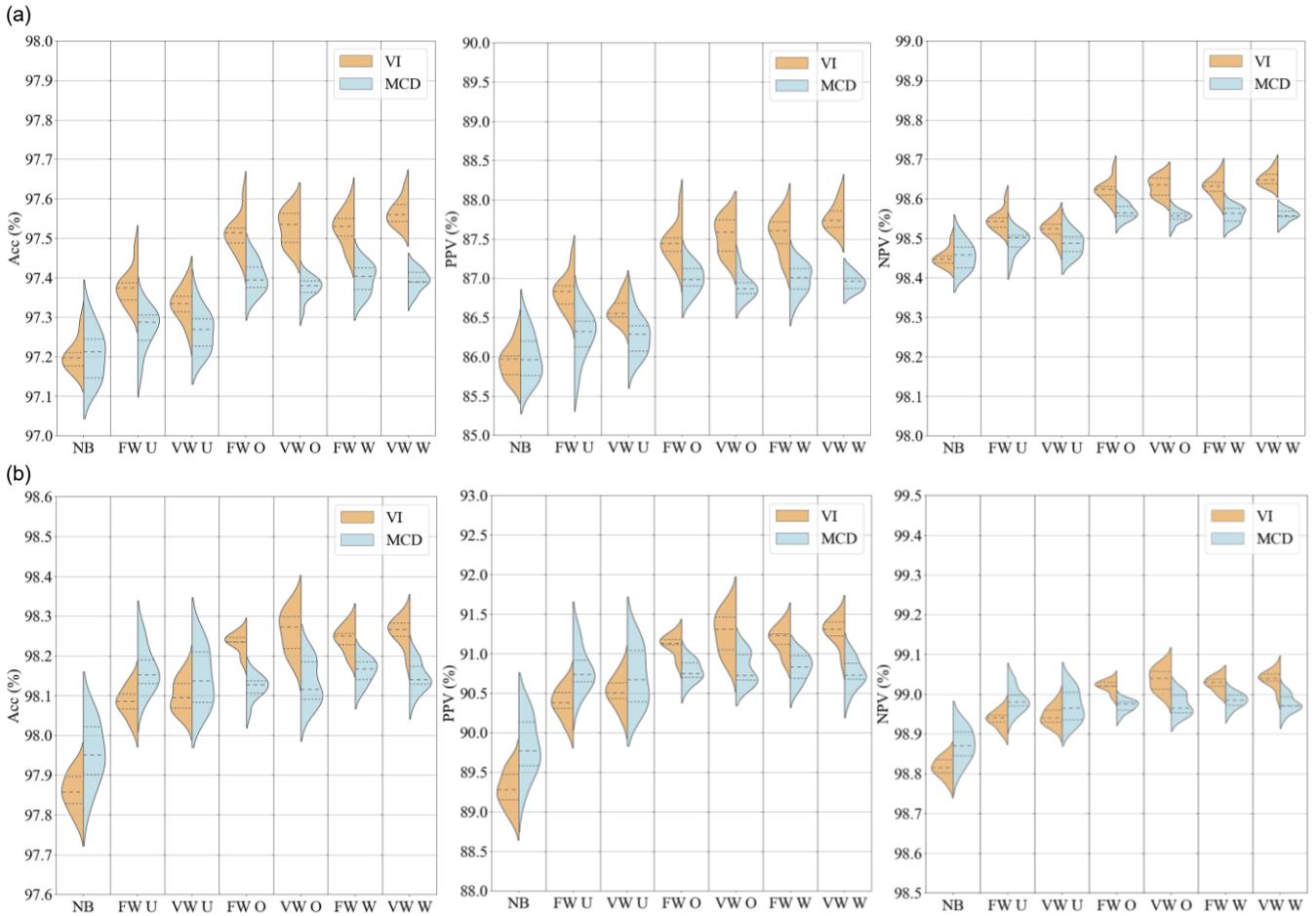

Fig. 4. Performance of the models, depicting the three quartiles, when *ProBoost* is not used (NB) and when a *ProBoost* variant (U – undersampled, O – oversampled, W – weighted) with four levels is used, examining (a) the first and (b) the second prepared datasets.

When comparing the results of *ProBoost* against those of the deterministic model (average Acc, PPV, and NPV of 97.2%, 86.4%, and 98.5%, respectively), we observe that even two levels of boosting (with either *ProBoost* methods) allow the probabilistic models (both VI-based and MCD-based) to surpass the deterministic model. These results clearly show that *ProBoost* can improve the performance of probabilistic models to exceed that of the standard deterministic models.

When using AdaBoost.M1 with four levels, the MCD model with FW attained an Acc, PPV, and NPV of 97.5%, 87.2%, and 98.6%, respectively. This performance is analogous to the equivalent model using *weighted ProBoost*. However, the AdaBoost.M1 VI model with FW reached an Acc, PPV, and NPV of 97.3%, 86.7%, and 98.5%, respectively. This performance is significantly lower than the equivalent model using weighted *ProBoost* (Acc, PPV, and NPV of 97.5%, 87.6%, and 98.6%, respectively). These results further support the relevance of using *ProBoost*, especially with VI.

The ROI was assessed for all performance metrics. It was observed that it yields the same values for Acc, Sen, Spe, and AUC, as expected, as the dataset is balanced. The plots presenting the ROI for the Acc of the VI and MCD models are shown in Fig. 5. Regarding ROI estimated for PPV and NPV, when compared to the ROI of the other performance metrics, the average difference (in absolute values) was lower than 0.25%. Hence the plots will be very similar to those presented in Fig. 5. By examining the figures, it is possible to support further the conclusion of the previous analysis (from Fig. 4 (a) and Tables 1, 2, and 3 in the supplementary material), as it is clear the improvement in the ROI attained by both *oversampled* and *weighted ProBoost* when the number of levels increases. It is also possible to observe that *undersampled ProBoost* is more suitable for an ensemble with just two weak learners.

### 5.4 Dataset with Additional Noise

The second prepared dataset was examined by a deterministic model and by the probabilistic models (VI and MCD models), using *ProBoost* with four levels (the top performer in the previous test) and without *ProBoost* (single classifier). The Acc, PPV, and NPV obtained are shown in Fig. 4 (b), while the full results are presented in Table 4 of the supplementary material. These results allow concluding that FW and VW reach similar performance. Regarding *undersampled ProBoost*, MCD outperforms VI, whereas the opposite is true for both *oversampled* and *weighted ProBoost*. Another relevant fact is the significance of the results, with p-value less than 0.05 for all examined performance metrics, emphasizing the relevance of the proposed methodology. Moreover, similarly to the previous test, the use of *ProBoost* allowed the probabilistic models to outperform the deterministic one.

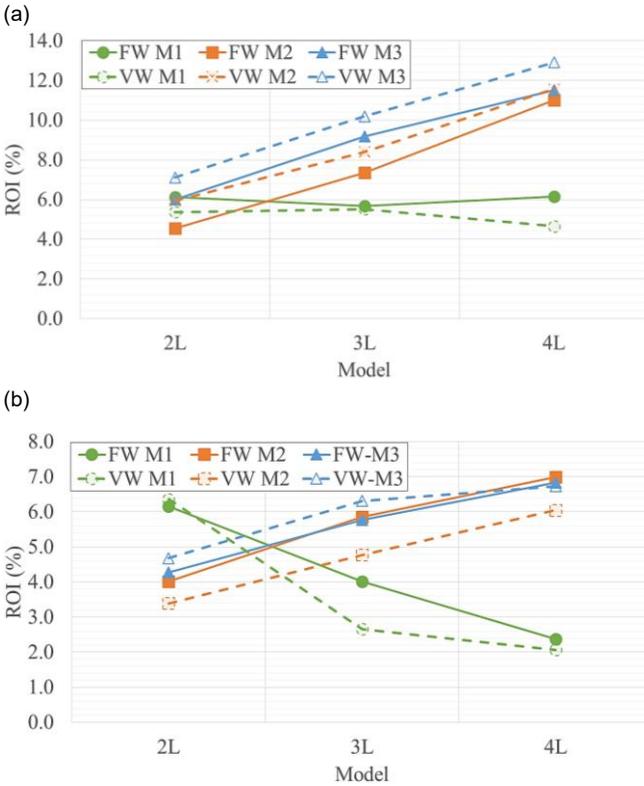

Fig. 5. ROI values for the (a) VI and (b) MCD models, using undersampled (M1), oversampled (M2), and weighted (M3) *ProBoost*, examining two (2L), three (3L), and four (4L) levels for each.

Fig. 6 presents the ROI of the studied VI and MCD models (which was the same for the Acc, Sen, Spe, and AUC). The average difference (in absolute values) for the ROI estimated for PPV and NPV was lower than 0.24% and 0.68% for VI and MCD, respectively. Analysis of Fig. 6 provides further support to the conclusions (Fig. 4 and Tables 1 to 4 in the supplementary material), as ROI denotes an improvement reached by using *ProBoost*. It is noticeable that *undersampled ProBoost* produced the highest ROI for the MCD-based model, while both oversampled and weighted *ProBoost* have the best ROI for the VI-based model with analogous results.

### 5.5 Analysis of the Ensemble Weights

To check the effect of the approaches used to set the ensemble weights, the best method from the previous analysis with *weighted Proboost* was used, with the number of levels increased to 10. The goal of this test is to compare the FW and VW approaches against a clairvoyant VW model in which weights are selected to maximize the performance on the test dataset, representing the best possible weights (this model is denoted as VWO). The goal is to check if the approaches used to determine the ensemble weights are too far from the best possible choice. For this purpose, the first prepared dataset was used. The best weights for the VWO ensemble model were found using MC sampling, running $10^4$ samples with the weight of each level randomly selected in [0, 1], and then checking the performance of each weight combination on the test dataset.

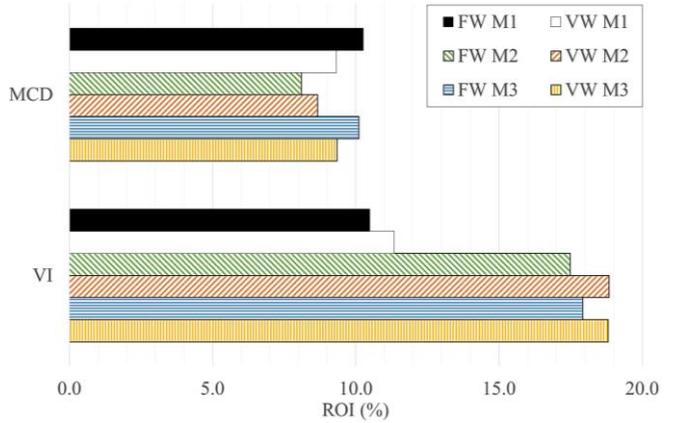

Fig. 6. ROI values for the models that analyzed the second prepared dataset, using undersampled (M1), oversampled (M2), and weighted (M3) *ProBoost*, all with four levels.

The goal is to determine the ROI improvement from no boosting to *ProBoost* with four levels, then the ROI improvement from *ProBoost* with four to ten levels. The attained results for AUC are shown in Fig. 7, depicting the ROI on top of the bars. Table 5 in the supplementary material has the results for the other performance metrics. The obtained results allow concluding that the MCD performance improvement when increasing the number of levels is low, especially for the VW approach. It is also possible to observe that FW leads to better results when the MCD-based model is used. The opposite was true for the VI models, though the performance difference was small. Also, for the VI-based model with ten levels, the ROI of the VWO approach was only 2% when compared with the VW model, and the tendency of saturation is less apparent, although it is present. The MCD-based model attained a similar result. It is, therefore, possible to conclude that the ensemble employed approaches are not excessively far from an optimal scenario, further corroborating the analysis.

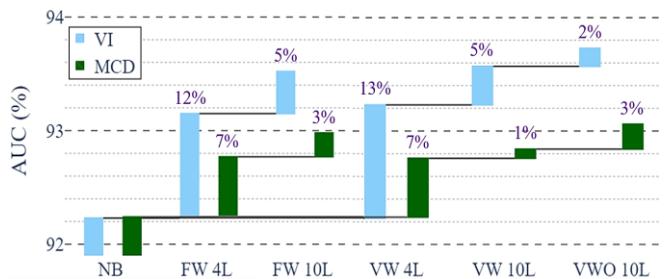

Fig. 7. AUC and ROI (shown above the bar) variations as the number of *ProBoost* levels (L) is increased, using weighted *ProBoost*.

To further check the performance saturation as the number of levels increases, the ROI results (compared with the performance without using *ProBoost*) for the AUC (the other performance metrics have a similar ROI) are presented in Fig. 8. The tendency lines are logarithmic, pointing towards saturation. It is, therefore, noticeable that although increasing the number of levels improves the performance, it shows a diminishing return behavior.

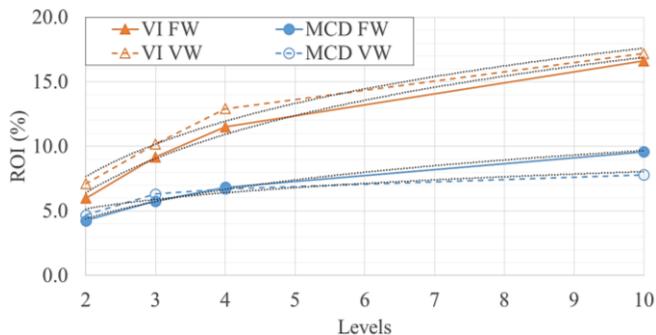

Fig. 8. ROI of the AUC values when the number of levels is increased, using weighted *ProBoost*.

## 6. Conclusion

This article introduced *ProBoost*, a boosting algorithm for probabilistic models. The proposed algorithm uses the sample-wise epistemic uncertainty estimates, an uncertainty metric that is accessible in probabilistic models, to progressively produce weak learners more specialized in the most challenging data samples (i.e., those with the highest uncertainty), thus implementing a boosting strategy. This algorithm is independent of which probabilistic classifier (weak learner) is used, only requiring the uncertainty assessments from each classifier.

The proposed model was instantiated using CNNs as weak learners. Although this work can be seen as presenting initial results, it is evident that the algorithm is promising. Nevertheless, several aspects have room for improvement, particularly in the technique adopted for using the ensemble. Another aspect is that *ProBoost* depends solely on the epistemic uncertainty assessment of each weak learner. However, it is likely that the inclusion of other factors, such as the performance of each weak learner, could further improve the robustness of the model and prevent performance deterioration. Furthermore, the hypothesis combination rules for producing the ensemble are suboptimal, and others should be studied. Therefore, the next steps in this research must further validate the proposed algorithm, check the results on other standard datasets, and test different classifiers as weak learners.

## Acknowledgment

This research was funded by LARSyS (Project - UIDB/50009/2020).

The authors acknowledge the Portuguese Foundation for Science and Technology (FCT) for support through Projeto Estratégico LA 9 – UID/EEA/50009/2019, and ARDITI (Agência Regional para o Desenvolvimento da Investigação, Tecnologia e Inovação) under the scope of the project M1420-09-5369-FSE-000002 – Post-Doctoral Fellowship, co-financed by the Madeira 14-20 Program - European Social Fund.

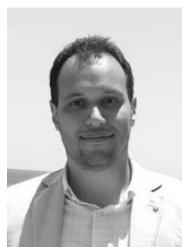

**Fábio Mendonça** received the BS and MSc degrees in electrical and telecommunications engineering from University of Madeira, and the PhD degree in electrical and computer engineering from Instituto Superior Técnico - University of Lisbon in partnership with Carnegie Mellon University. He works at the University of Madeira and is a researcher with the Interactive Technologies Institute - LARSyS. His research interests include sleep analysis, pattern recognition, and machine learning.

13
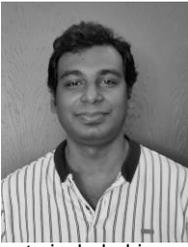

**Sheikh Shanawaz Mostafa** received a Ph.D. degree from Instituto Superior Técnico, of the University of Lisbon, Portugal, in 2020. Previously received the M.S. degree from Khulna University of Engineering & Technology, Bangladesh, in 2012 and B.S. Engg. degree from the University of Khulna, Bangladesh, in 2010. He worked on different research projects related to biomedical signal and image processing and hardware implementation. His research interests include biomedical signal processing, artificial neural networks, and hardware implementations.

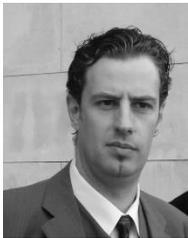

**Fernando Morgado-Dias** received the B.S. degree from the University of Aveiro, Portugal, in 1994, the M.S. degree from University Joseph Fourier, Grenoble, France, in 1995, and the Ph.D. degree from the University of Aveiro in 2005. He was a Lecturer with the Technical University of Setúbal, Portugal. He is currently Associate Professor with the Universidade da Madeira and Researcher with the ITI/Larsys and Madeira Interactive Technologies Institute. His research interests include sleep monitoring, renewable energy, artificial neural networks and FPGA implementations. Dr. Morgado-Dias was previously President and Vice-President of the Portuguese Association of Automatic Control and is Director of PhD degree in Electrical Engineering.

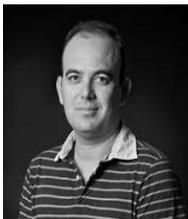

**Antonio G. Ravelo-García** received the M.S. and Ph.D. degrees in Telecommunication Engineering from Universidad de Las Palmas de Gran Canaria, Spain. He is Associate Professor in the Department of Signal and Communications (DSC) and Institute for Technological Development and Innovation in Communications (IDeTIC) at University of Las Palmas de Gran Canaria (ULPGC), Spain. He is also an international collaborator member of ITI/LARSyS (Portugal). He has participated in multiple research projects and publications. His current research interest includes biomedical signal processing, nonlinear signal analysis, data mining and intelligent systems in multimedia, communications, environment, health and wellbeing.

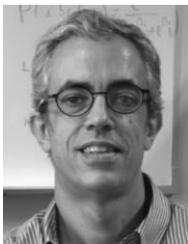

**Mário A. T. Figueiredo** received the E.E., M.Sc., Ph.D., and Agregado degrees in electrical and computer engineering from Instituto Superior Tecnico (IST), University of Lisbon, Lisbon, Portugal, in 1985, 1990, 1994, and 2004, respectively. He has been a faculty member with the Department of Electrical and Computer Engineering, IST, where he is a Distinguished Professor. He is also a Group and Area Coordinator with the Instituto de Telecomunicações. His research interests include signal processing, machine learning, and optimization. He is a fellow of the International Association for Pattern Recognition (IAPR), of the IEEE, and of the European Association for Signal Processing (EURASIP). He is a member of the Portuguese Academy of Engineering and the Lisbon Academy of Sciences. He received several major awards, namely the 2011 IEEE Signal Processing Society Best Paper Award, the 2014 IEEE W. R. G. Baker Award, the 2016 EURASIP Individual Technical Achievement Award, and the 2016 IAPR Pierre Devijver Award. He has been an Associate Editor of many journals. He has been included in the Highly Cited Researchers list (Clarivate Analytics, formerly by Thomson Reuters) from 2014 to 2018.

# Supplementary material

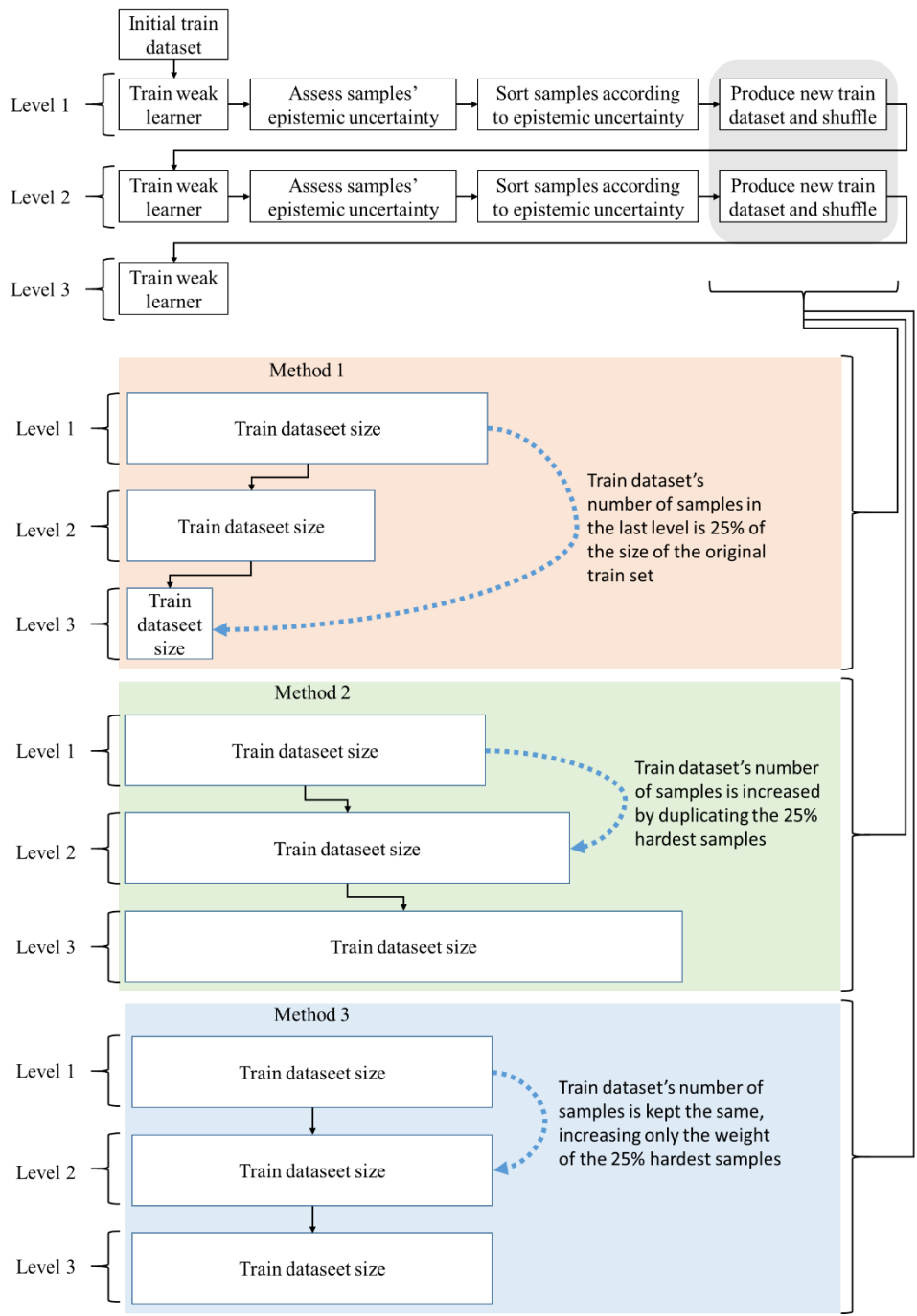

**Fig. 1.** Example of ProBoost for a model with three levels.

TABLE 1
PERFORMANCE OF THE VI MODELS WITH ONE, TWO, THREE, AND FOUR LEVELS. NOTE: ONLY THE P-VALUE IS NOT PRESENTED IN PERCENTAGE

| V | | Acc (%) | | Sen (%) | | Spe (%) | | PPV (%) | | NPV (%) | | AUC (%) | |
|---|---|---|---|---|---|---|---|---|---|---|---|---|---|
| 1 | μ | 97.20 | | 86.02 | | 98.45 | | 85.94 | | 98.45 | | 92.23 | |
|   | σ | 0.04 | | 0.19 | | 0.02 | | 0.20 | | 0.02 | | 0.10 | |
|   | ∧ | 97.16 | | 85.81 | | 98.42 | | 85.68 | | 98.43 | | 92.12 | |
|   | ∨ | 97.29 | | 86.45 | | 98.49 | | 86.34 | | 98.50 | | 92.47 | |
|   | CI | 97.18, 97.23 | | 85.90, 86.13 | | 98.43, 98.46 | | 86.44, 86.06 | | 98.44, 98.46 | | 92.17, 92.30 | |
| **Undersampled ProBoost (method one)** | | | | | | | | | | | | | |
| | | FW | VW | FW | VW | FW | VW | FW | VW | FW | VW | FW | VW |
| 2 | μ | 97.37 | 97.35 | 86.87 | 86.77 | 98.54 | 98.53 | 86.80 | 86.71 | 98.55 | 98.53 | 92.71 | 92.65 |
|   | σ | 0.04 | 0.05 | 0.22 | 0.23 | 0.02 | 0.03 | 0.20 | 0.22 | 0.02 | 0.03 | 0.12 | 0.13 |
|   | ∧ | 97.29 | 97.29 | 86.46 | 86.43 | 98.50 | 98.49 | 86.41 | 86.39 | 98.50 | 98.50 | 92.48 | 92.46 |
|   | ∨ | 97.43 | 97.44 | 87.17 | 87.22 | 98.57 | 98.58 | 87.10 | 87.15 | 98.58 | 98.59 | 92.87 | 92.90 |
|   | CI | 97.35, 97.40 | 97.32, 97.38 | 86.74, 87.00 | 86.62, 86.91 | 98.53, 98.56 | 98.51, 98.55 | 86.67, 86.92 | 86.58, 86.85 | 98.53, 98.56 | 98.52, 98.55 | 92.63, 92.78 | 92.57, 92.73 |
|   | p | < 10⁻³ | < 10⁻³ | < 10⁻³ | < 10⁻³ | < 10⁻³ | < 10⁻³ | < 10⁻³ | < 10⁻³ | < 10⁻³ | < 10⁻³ | < 10⁻³ | < 10⁻³ |
| 3 | μ | 97.36 | 97.36 | 86.81 | 86.79 | 98.53 | 98.53 | 86.74 | 86.74 | 98.54 | 98.54 | 92.67 | 92.66 |
|   | σ | 0.04 | 0.06 | 0.18 | 0.30 | 0.02 | 0.03 | 0.19 | 0.30 | 0.02 | 0.03 | 0.10 | 0.16 |
|   | ∧ | 97.32 | 97.29 | 86.62 | 86.43 | 98.51 | 98.49 | 86.53 | 86.30 | 98.52 | 98.50 | 92.57 | 92.46 |
|   | ∨ | 97.43 | 97.50 | 87.17 | 87.50 | 98.57 | 98.61 | 87.12 | 87.43 | 98.58 | 98.62 | 92.87 | 93.06 |
|   | CI | 97.34, 97.38 | 97.32, 97.39 | 86.70, 86.92 | 86.61, 86.97 | 98.52, 98.55 | 98.51, 98.55 | 86.62, 86.85 | 86.55, 86.92 | 98.53, 98.55 | 98.52, 98.56 | 92.61, 92.74 | 92.56, 92.76 |
|   | p | < 10⁻³ | < 10⁻³ | < 10⁻³ | < 10⁻³ | < 10⁻³ | < 10⁻³ | < 10⁻³ | < 10⁻³ | < 10⁻³ | < 10⁻³ | < 10⁻³ | < 10⁻³ |
| 4 | μ | 97.38 | 97.33 | 86.88 | 86.67 | 98.54 | 98.52 | 86.83 | 86.60 | 98.55 | 98.52 | 92.71 | 92.59 |
|   | σ | 0.04 | 0.04 | 0.21 | 0.20 | 0.02 | 0.02 | 0.20 | 0.15 | 0.02 | 0.02 | 0.12 | 0.11 |
|   | ∧ | 97.32 | 97.26 | 86.59 | 86.28 | 98.51 | 98.48 | 86.51 | 86.37 | 98.51 | 98.48 | 92.55 | 92.38 |
|   | ∨ | 97.48 | 97.40 | 87.38 | 87.01 | 98.60 | 98.56 | 87.28 | 86.90 | 98.60 | 98.56 | 92.99 | 92.78 |
|   | CI | 97.35, 97.40 | 97.31, 97.36 | 86.74, 87.01 | 86.55, 86.79 | 98.53, 98.56 | 98.51, 98.53 | 86.70, 86.95 | 86.51, 86.69 | 98.53, 98.56 | 98.51, 98.54 | 92.64, 92.78 | 92.53, 92.66 |
|   | p | < 10⁻³ | < 10⁻³ | < 10⁻³ | < 10⁻³ | < 10⁻³ | < 10⁻³ | < 10⁻³ | < 10⁻³ | < 10⁻³ | < 10⁻³ | < 10⁻³ | < 10⁻³ |
| **Oversampled ProBoost (method two)** | | | | | | | | | | | | | |
| | | FW | VW | FW | VW | FW | VW | FW | VW | FW | VW | FW | VW |
| 2 | μ | 97.33 | 97.37 | 86.65 | 86.85 | 98.52 | 98.54 | 86.53 | 86.70 | 98.52 | 98.55 | 92.58 | 92.69 |
|   | σ | 0.04 | 0.04 | 0.19 | 0.18 | 0.02 | 0.02 | 0.21 | 0.18 | 0.02 | 0.02 | 0.11 | 0.10 |
|   | ∧ | 97.27 | 97.33 | 86.34 | 86.64 | 98.48 | 98.52 | 86.17 | 86.47 | 98.49 | 98.52 | 92.41 | 92.58 |
|   | ∨ | 97.39 | 97.42 | 86.95 | 87.11 | 98.55 | 98.57 | 86.86 | 86.98 | 98.56 | 98.57 | 92.75 | 92.84 |
|   | CI | 97.31, 97.35 | 97.35, 97.39 | 86.53, 86.77 | 86.74, 86.96 | 98.50, 98.53 | 98.53, 98.55 | 86.40, 86.66 | 86.59, 86.81 | 98.51, 98.54 | 98.53, 98.56 | 92.52, 92.65 | 92.63, 92.75 |
|   | p | < 10⁻³ | < 10⁻³ | < 10⁻³ | < 10⁻³ | < 10⁻³ | < 10⁻³ | < 10⁻³ | < 10⁻³ | < 10⁻³ | < 10⁻³ | < 10⁻³ | < 10⁻³ |
| 3 | μ | 97.41 | 97.44 | 87.05 | 87.19 | 98.56 | 98.58 | 86.94 | 87.08 | 98.57 | 98.58 | 92.80 | 92.88 |
|   | σ | 0.02 | 0.04 | 0.12 | 0.19 | 0.01 | 0.02 | 0.12 | 0.21 | 0.01 | 0.02 | 0.07 | 0.10 |
|   | ∧ | 97.37 | 97.35 | 86.84 | 86.75 | 98.54 | 98.53 | 86.73 | 86.62 | 98.54 | 98.54 | 92.69 | 92.64 |
|   | ∨ | 97.44 | 97.49 | 87.20 | 87.45 | 98.58 | 98.61 | 87.13 | 87.38 | 98.58 | 98.61 | 92.89 | 93.03 |
|   | CI | 97.39, 97.42 | 97.41, 97.46 | 86.97, 87.12 | 87.07, 87.31 | 98.55, 98.57 | 98.56, 98.59 | 86.87, 87.02 | 86.96, 87.21 | 98.56, 98.57 | 98.57, 98.59 | 92.76, 92.85 | 92.82, 92.95 |
|   | p | < 10⁻³ | < 10⁻³ | < 10⁻³ | < 10⁻³ | < 10⁻³ | < 10⁻³ | < 10⁻³ | < 10⁻³ | < 10⁻³ | < 10⁻³ | < 10⁻³ | < 10⁻³ |
| 4 | μ | 97.51 | 97.53 | 87.56 | 87.63 | 98.62 | 98.63 | 87.45 | 87.54 | 98.62 | 98.63 | 93.09 | 93.13 |
|   | σ | 0.04 | 0.04 | 0.22 | 0.21 | 0.02 | 0.02 | 0.22 | 0.22 | 0.02 | 0.02 | 0.12 | 0.11 |
|   | ∧ | 97.44 | 97.47 | 87.22 | 87.34 | 98.58 | 98.59 | 87.11 | 87.21 | 98.58 | 98.60 | 92.90 | 92.97 |
|   | ∨ | 97.61 | 97.59 | 88.06 | 87.93 | 98.67 | 98.66 | 87.97 | 87.82 | 98.68 | 98.66 | 93.37 | 93.29 |
|   | CI | 97.48, 97.54 | 97.50, 97.55 | 87.42, 87.69 | 87.50, 87.76 | 98.60, 98.63 | 98.61, 98.64 | 87.31, 87.58 | 87.41, 87.67 | 98.61, 98.64 | 98.62, 98.65 | 93.01, 93.16 | 93.06, 93.20 |
|   | p | < 10⁻³ | < 10⁻³ | < 10⁻³ | < 10⁻³ | < 10⁻³ | < 10⁻³ | < 10⁻³ | < 10⁻³ | < 10⁻³ | < 10⁻³ | < 10⁻³ | < 10⁻³ |
| **Weighted ProBoost (method three)** | | | | | | | | | | | | | |

|  |  | FW | VW | FW | VW | FW | VW | FW | VW | FW | VW | FW | VW |
|---|---|---|---|---|---|---|---|---|---|---|---|---|---|
| 2 | μ | 97.37 | 97.40 | 86.85 | 87.01 | 98.54 | 98.56 | 86.76 | 86.99 | 98.54 | 98.56 | 92.70 | 92.78 |
|  | σ | 0.03 | 0.03 | 0.16 | 0.16 | 0.02 | 0.02 | 0.15 | 0.20 | 0.02 | 0.02 | 0.09 | 0.09 |
|  | ∧ | 97.31 | 97.35 | 86.57 | 86.77 | 98.51 | 98.53 | 86.51 | 86.68 | 98.51 | 98.54 | 92.54 | 92.65 |
|  | ∨ | 97.42 | 97.46 | 87.12 | 87.29 | 98.57 | 98.59 | 87.01 | 87.33 | 98.57 | 98.59 | 92.84 | 92.94 |
|  | CI | 97.35, 97.39 | 97.38, 97.42 | 86.75, 86.96 | 86.91, 87.11 | 98.53, 98.55 | 98.55, 98.57 | 86.67, 86.86 | 86.86, 87.11 | 98.53, 98.56 | 98.55, 98.57 | 92.64, 92.75 | 92.73, 92.84 |
|  | p | < 10⁻³ | < 10⁻³ | < 10⁻³ | < 10⁻³ | < 10⁻³ | < 10⁻³ | < 10⁻³ | < 10⁻³ | < 10⁻³ | < 10⁻³ | < 10⁻³ | < 10⁻³ |
| 3 | μ | 97.46 | 97.49 | 87.30 | 87.44 | 98.59 | 98.60 | 87.23 | 87.41 | 98.59 | 98.61 | 92.94 | 93.02 |
|  | σ | 0.03 | 0.04 | 0.16 | 0.20 | 0.02 | 0.02 | 0.18 | 0.21 | 0.02 | 0.02 | 0.09 | 0.11 |
|  | ∧ | 97.41 | 97.43 | 87.07 | 87.17 | 98.56 | 98.57 | 86.94 | 87.07 | 98.57 | 98.58 | 92.82 | 92.87 |
|  | ∨ | 97.51 | 97.58 | 87.57 | 87.89 | 98.62 | 98.65 | 87.51 | 87.85 | 98.62 | 98.66 | 93.09 | 93.27 |
|  | CI | 97.44, 97.48 | 97.46, 97.51 | 87.20, 87.40 | 87.32, 87.56 | 98.58, 98.60 | 98.59, 98.62 | 87.12, 87.34 | 87.28, 87.54 | 98.58, 98.60 | 98.60, 98.62 | 92.89, 93.00 | 92.96, 93.09 |
|  | p | < 10⁻³ | < 10⁻³ | < 10⁻³ | < 10⁻³ | < 10⁻³ | < 10⁻³ | < 10⁻³ | < 10⁻³ | < 10⁻³ | < 10⁻³ | < 10⁻³ | < 10⁻³ |
| 4 | μ | 97.52 | 97.56 | 87.62 | 87.82 | 98.62 | 98.65 | 87.56 | 87.77 | 98.63 | 98.65 | 93.12 | 93.23 |
|  | σ | 0.04 | 0.03 | 0.21 | 0.16 | 0.02 | 0.02 | 0.23 | 0.16 | 0.02 | 0.02 | 0.12 | 0.09 |
|  | ∧ | 97.46 | 97.52 | 87.28 | 87.61 | 98.59 | 98.62 | 87.18 | 87.55 | 98.59 | 98.63 | 92.93 | 93.12 |
|  | ∨ | 97.60 | 97.63 | 87.99 | 88.16 | 98.67 | 98.68 | 87.91 | 88.11 | 98.67 | 98.69 | 93.33 | 93.42 |
|  | CI | 97.50, 97.55 | 97.54, 97.58 | 87.49, 87.75 | 87.72, 87.92 | 98.61, 98.64 | 98.64, 98.66 | 87.42, 87.70 | 87.67, 87.87 | 98.62, 98.64 | 98.64, 98.66 | 93.05, 93.20 | 93.18, 93.29 |
|  | p | < 10⁻³ | < 10⁻³ | < 10⁻³ | < 10⁻³ | < 10⁻³ | < 10⁻³ | < 10⁻³ | < 10⁻³ | < 10⁻³ | < 10⁻³ | < 10⁻³ | < 10⁻³ |

TABLE 2
PERFORMANCE OF THE MCD MODELS WITH ONE, TWO, THREE, AND FOUR LEVELS. NOTE: ONLY THE P-VALUE IS NOT PRESENTED IN PERCENTAGE

| V |  | Acc (%) | | Sen (%) | | Spe (%) | | PPV (%) | | NPV (%) | | AUC (%) | |
|---|---|---|---|---|---|---|---|---|---|---|---|---|---|
| 1 | μ | 97.21 | | 86.03 | | 98.45 | | 86.00 | | 98.46 | | 92.24 | |
|  | σ | 0.06 | | 0.30 | | 0.03 | | 0.27 | | 0.03 | | 0.17 | |
|  | ∧ | 97.12 | | 85.62 | | 98.40 | | 85.64 | | 98.41 | | 92.01 | |
|  | ∨ | 97.31 | | 86.57 | | 98.51 | | 86.50 | | 98.52 | | 92.54 | |
|  | CI | 97.17, 97.24 | | 85.84, 86.22 | | 98.43, 98.47 | | 85.83, 86.17 | | 98.44, 98.48 | | 92.13, 92.34 | |
| **Undersampled ProBoost (method one)** | | | | | | | | | | | | | |
|  |  | FW | VW | FW | VW | FW | VW | FW | VW | FW | VW | FW | VW |
| 2 | μ | 97.38 | 97.38 | 86.89 | 86.92 | 98.54 | 98.55 | 86.82 | 86.83 | 98.55 | 98.55 | 92.72 | 92.73 |
|  | σ | 0.06 | 0.04 | 0.29 | 0.21 | 0.03 | 0.02 | 0.30 | 0.22 | 0.03 | 0.02 | 0.16 | 0.11 |
|  | ∧ | 97.28 | 97.30 | 86.41 | 86.48 | 98.49 | 98.50 | 86.30 | 86.39 | 98.50 | 98.50 | 92.45 | 92.49 |
|  | ∨ | 97.46 | 97.44 | 87.31 | 87.21 | 98.59 | 98.58 | 87.30 | 87.17 | 98.59 | 98.58 | 92.95 | 92.89 |
|  | CI | 97.34, 97.41 | 97.36, 97.41 | 86.71, 87.07 | 86.79, 87.05 | 98.52, 98.56 | 98.53, 98.56 | 86.64, 87.01 | 86.69, 86.96 | 98.53, 98.57 | 98.54, 98.57 | 92.62, 92.82 | 92.66, 92.80 |
|  | p | < 10⁻³ | < 10⁻³ | < 10⁻³ | < 10⁻³ | < 10⁻³ | < 10⁻³ | < 10⁻³ | < 10⁻³ | < 10⁻³ | < 10⁻³ | < 10⁻³ | < 10⁻³ |
| 3 | μ | 97.32 | 97.28 | 86.59 | 86.40 | 98.51 | 98.49 | 86.54 | 86.33 | 98.52 | 98.50 | 92.55 | 92.44 |
|  | σ | 0.04 | 0.06 | 0.18 | 0.28 | 0.02 | 0.03 | 0.18 | 0.28 | 0.02 | 0.03 | 0.10 | 0.15 |
|  | ∧ | 97.27 | 97.19 | 86.37 | 85.97 | 98.49 | 98.44 | 86.25 | 85.90 | 98.49 | 98.45 | 92.43 | 92.21 |
|  | ∨ | 97.38 | 97.40 | 86.91 | 87.01 | 98.55 | 98.56 | 86.81 | 86.89 | 98.55 | 98.56 | 92.73 | 92.78 |
|  | CI | 97.30, 97.34 | 97.25, 97.31 | 86.48, 86.70 | 86.23, 86.57 | 98.50, 98.52 | 98.47, 98.51 | 86.43, 86.65 | 86.16, 86.50 | 98.50, 98.53 | 98.48, 98.52 | 92.49, 92.61 | 92.35, 92.54 |
|  | p | < 10⁻³ | < 10⁻³ | < 10⁻³ | < 10⁻³ | < 10⁻³ | < 10⁻³ | < 10⁻³ | < 10⁻³ | < 10⁻³ | < 10⁻³ | < 10⁻³ | < 10⁻³ |
| 4 | μ | 97.27 | 97.26 | 86.36 | 86.32 | 98.48 | 98.48 | 86.28 | 86.26 | 98.49 | 98.49 | 92.42 | 92.40 |
|  | σ | 0.05 | 0.05 | 0.24 | 0.25 | 0.03 | 0.03 | 0.28 | 0.23 | 0.03 | 0.03 | 0.13 | 0.14 |
|  | ∧ | 97.16 | 97.20 | 85.81 | 85.98 | 98.42 | 98.44 | 85.67 | 85.91 | 98.43 | 98.45 | 92.12 | 92.21 |
|  | ∨ | 97.34 | 97.36 | 86.69 | 86.78 | 98.52 | 98.53 | 86.67 | 86.74 | 98.53 | 98.54 | 92.61 | 92.66 |
|  | CI | 97.24, 97.30 | 97.23, 97.29 | 86.21, 86.51 | 86.17, 86.47 | 98.47, 98.50 | 98.46, 98.50 | 86.11, 86.45 | 86.12, 86.41 | 98.48, 98.51 | 98.47, 98.50 | 92.34, 92.51 | 92.31, 92.48 |
|  | p | **0.01** | **0.02** | **0.01** | **0.02** | **0.01** | **0.02** | **0.02** | **0.02** | **0.01** | **0.03** | **0.01** | **0.02** |
| **Oversampled ProBoost (method two)** | | | | | | | | | | | | | |

|   |   | FW | VW | FW | VW | FW | VW | FW | VW | FW | VW | FW | VW |
|---|---|---|---|---|---|---|---|---|---|---|---|---|---|
| 2 | μ | 97.32 | 97.30 | 86.59 | 86.50 | 98.51 | 98.50 | 86.54 | 86.47 | 98.52 | 98.51 | 92.55 | 92.50 |
|   | σ | 0.04 | 0.04 | 0.19 | 0.22 | 0.02 | 0.02 | 0.19 | 0.19 | 0.02 | 0.02 | 0.10 | 0.12 |
|   | ∧ | 97.26 | 97.22 | 86.28 | 86.10 | 98.48 | 98.46 | 86.24 | 86.04 | 98.48 | 98.47 | 92.38 | 92.28 |
|   | ∨ | 97.36 | 97.37 | 86.81 | 86.83 | 98.53 | 98.54 | 86.79 | 86.70 | 98.55 | 98.55 | 92.67 | 92.68 |
|   | CI | 97.30, 97.34 | 97.27, 97.33 | 86.48, 86.71 | 86.37, 86.64 | 98.50, 98.52 | 98.49, 98.52 | 86.42, 86.66 | 86.35, 86.59 | 98.51, 98.53 | 98.50, 98.53 | 92.49, 92.61 | 92.43, 92.58 |
|   | p | $< 10^{-3}$ | $< 10^{-3}$ | $< 10^{-3}$ | $< 10^{-3}$ | $< 10^{-3}$ | $< 10^{-3}$ | $< 10^{-3}$ | $< 10^{-3}$ | $< 10^{-3}$ | $< 10^{-3}$ | $< 10^{-3}$ | $< 10^{-3}$ |
| 3 | μ | 97.37 | 97.34 | 86.85 | 86.70 | 98.54 | 98.52 | 86.83 | 86.71 | 98.55 | 98.54 | 92.69 | 92.61 |
|   | σ | 0.03 | 0.04 | 0.17 | 0.21 | 0.02 | 0.02 | 0.20 | 0.21 | 0.02 | 0.02 | 0.09 | 0.12 |
|   | ∧ | 97.32 | 97.24 | 86.62 | 86.22 | 98.51 | 98.47 | 86.61 | 86.28 | 98.53 | 98.49 | 92.57 | 92.34 |
|   | ∨ | 97.43 | 97.38 | 87.13 | 86.92 | 98.57 | 98.55 | 87.30 | 87.12 | 98.59 | 98.56 | 92.85 | 92.73 |
|   | CI | 97.35, 97.39 | 97.31, 97.37 | 86.74, 86.95 | 86.57, 86.83 | 98.53, 98.55 | 98.51, 98.54 | 86.71, 86.96 | 86.58, 86.85 | 98.54, 98.56 | 98.52, 98.55 | 92.64, 92.75 | 92.54, 92.68 |
|   | p | $< 10^{-3}$ | $< 10^{-3}$ | $< 10^{-3}$ | $< 10^{-3}$ | $< 10^{-3}$ | $< 10^{-3}$ | $< 10^{-3}$ | $< 10^{-3}$ | $< 10^{-3}$ | $< 10^{-3}$ | $< 10^{-3}$ | $< 10^{-3}$ |
| 4 | μ | 97.40 | 97.38 | 87.01 | 86.88 | 98.56 | 98.54 | 87.00 | 86.89 | 98.57 | 98.56 | 92.78 | 92.71 |
|   | σ | 0.04 | 0.03 | 0.19 | 0.13 | 0.02 | 0.01 | 0.17 | 0.15 | 0.02 | 0.01 | 0.10 | 0.07 |
|   | ∧ | 97.34 | 97.31 | 86.71 | 86.57 | 98.52 | 98.51 | 86.72 | 86.69 | 98.54 | 98.52 | 92.62 | 92.54 |
|   | ∨ | 97.46 | 97.41 | 87.31 | 87.04 | 98.59 | 98.56 | 87.27 | 87.23 | 98.60 | 98.58 | 92.95 | 92.80 |
|   | CI | 97.38, 97.42 | 97.36, 97.39 | 86.89, 87.12 | 86.79, 86.96 | 98.54, 98.57 | 98.53, 98.55 | 86.89, 87.10 | 86.80, 86.98 | 98.56, 98.58 | 98.55, 98.56 | 92.72, 92.85 | 92.66, 92.75 |
|   | p | $< 10^{-3}$ | $< 10^{-3}$ | $< 10^{-3}$ | $< 10^{-3}$ | $< 10^{-3}$ | $< 10^{-3}$ | $< 10^{-3}$ | $< 10^{-3}$ | $< 10^{-3}$ | $< 10^{-3}$ | $< 10^{-3}$ | $< 10^{-3}$ |
|   |   | **Weighted ProBoost (method three)** | | | | | | | | | | | |
|   |   | FW | VW | FW | VW | FW | VW | FW | VW | FW | VW | FW | VW |
| 2 | μ | 97.33 | 97.34 | 86.63 | 86.68 | 98.51 | 98.52 | 86.66 | 86.64 | 98.52 | 98.53 | 92.57 | 92.60 |
|   | σ | 0.07 | 0.02 | 0.35 | 0.11 | 0.04 | 0.01 | 0.33 | 0.10 | 0.04 | 0.01 | 0.20 | 0.06 |
|   | ∧ | 97.21 | 97.30 | 86.03 | 86.50 | 98.45 | 98.50 | 86.01 | 86.48 | 98.45 | 98.50 | 92.24 | 92.50 |
|   | ∨ | 97.42 | 97.37 | 87.10 | 86.85 | 98.57 | 98.54 | 87.07 | 86.81 | 98.57 | 98.55 | 92.83 | 92.69 |
|   | CI | 97.28, 97.37 | 97.32, 97.35 | 86.41, 86.84 | 86.62, 86.75 | 98.49, 98.54 | 98.51, 98.53 | 86.46, 86.86 | 86.58, 86.70 | 98.50, 98.54 | 98.52, 98.53 | 92.45, 92.69 | 92.57, 92.64 |
|   | p | $< 10^{-3}$ | $< 10^{-3}$ | $< 10^{-3}$ | $< 10^{-3}$ | $< 10^{-3}$ | $< 10^{-3}$ | $< 10^{-3}$ | $< 10^{-3}$ | $< 10^{-3}$ | $< 10^{-3}$ | $< 10^{-3}$ | $< 10^{-3}$ |
| 3 | μ | 97.37 | 97.38 | 86.84 | 86.91 | 98.54 | 98.55 | 86.85 | 86.87 | 98.54 | 98.55 | 92.69 | 92.73 |
|   | σ | 0.06 | 0.03 | 0.30 | 0.15 | 0.03 | 0.02 | 0.29 | 0.15 | 0.03 | 0.02 | 0.16 | 0.09 |
|   | ∧ | 97.27 | 97.33 | 86.37 | 86.64 | 98.49 | 98.52 | 86.25 | 86.56 | 98.49 | 98.52 | 92.43 | 92.58 |
|   | ∨ | 97.44 | 97.43 | 87.22 | 87.13 | 98.58 | 98.57 | 87.16 | 87.09 | 98.59 | 98.58 | 92.90 | 92.85 |
|   | CI | 97.33, 97.40 | 97.36, 97.40 | 86.65, 87.02 | 86.82, 87.01 | 98.52, 98.56 | 98.54, 98.56 | 86.67, 87.03 | 86.78, 86.96 | 98.52, 98.56 | 98.54, 98.56 | 92.58, 92.79 | 92.68, 92.78 |
|   | p | $< 10^{-3}$ | $< 10^{-3}$ | $< 10^{-3}$ | $< 10^{-3}$ | $< 10^{-3}$ | $< 10^{-3}$ | $< 10^{-3}$ | $< 10^{-3}$ | $< 10^{-3}$ | $< 10^{-3}$ | $< 10^{-3}$ | $< 10^{-3}$ |
| 4 | μ | 97.40 | 97.39 | 86.99 | 86.97 | 98.55 | 98.55 | 86.98 | 86.95 | 98.56 | 98.56 | 92.77 | 92.76 |
|   | σ | 0.04 | 0.02 | 0.18 | 0.12 | 0.02 | 0.01 | 0.18 | 0.10 | 0.02 | 0.01 | 0.10 | 0.07 |
|   | ∧ | 97.34 | 97.35 | 86.70 | 86.75 | 98.52 | 98.53 | 86.63 | 86.82 | 98.53 | 98.53 | 92.61 | 92.64 |
|   | ∨ | 97.45 | 97.43 | 87.26 | 87.15 | 98.58 | 98.57 | 87.24 | 87.12 | 98.59 | 98.58 | 92.92 | 92.86 |
|   | CI | 97.37, 97.42 | 97.38, 97.41 | 86.87, 87.10 | 86.89, 87.05 | 98.54, 98.57 | 98.54, 98.56 | 86.87, 87.09 | 86.89, 87.01 | 98.55, 98.57 | 98.55, 98.57 | 92.71, 92.83 | 92.72, 92.80 |
|   | p | $< 10^{-3}$ | $< 10^{-3}$ | $< 10^{-3}$ | $< 10^{-3}$ | $< 10^{-3}$ | $< 10^{-3}$ | $< 10^{-3}$ | $< 10^{-3}$ | $< 10^{-3}$ | $< 10^{-3}$ | $< 10^{-3}$ | $< 10^{-3}$ |

TABLE 3
PERFORMANCE OF THE DETERMINISTIC MODEL

|   | Acc (%) | Sen (%) | Spe (%) | PPV (%) | NPV (%) | AUC (%) |
|---|---|---|---|---|---|---|
| μ | 97.24 | 86.19 | 98.47 | 86.41 | 98.47 | 92.33 |
| σ | 0.05 | 0.23 | 0.03 | 0.30 | 0.03 | 0.13 |
| ∧ | 97.17 | 85.85 | 98.43 | 85.91 | 98.43 | 92.14 |
| ∨ | 97.31 | 86.57 | 98.51 | 86.84 | 98.51 | 92.54 |
| CI | 97.21, 97.27 | 86.04, 86.34 | 98.45, 98.48 | 86.22, 86.60 | 98.45, 98.48 | 92.25, 92.41 |

5TABLE 4
Performance of the deterministic, VI, and MCD models when examining the second prepared dataset. Note: only the p-value is not presented in percentage

| Model | | Acc (%) | | Sen (%) | | Spe (%) | | PPV (%) | | NPV (%) | | AUC (%) | |
|---|---|---|---|---|---|---|---|---|---|---|---|---|---|
| | | \multicolumn{12}{c}{Models without ProBoost} |
| Deterministic | μ | 98.11 | | 90.56 | | 98.95 | | 90.56 | | 98.96 | | 94.76 | |
| | σ | 0.08 | | 0.39 | | 0.04 | | 0.36 | | 0.04 | | 0.21 | |
| | ∧ | 97.99 | | 89.94 | | 98.88 | | 90.01 | | 98.89 | | 94.41 | |
| | ∨ | 98.25 | | 91.25 | | 99.03 | | 91.22 | | 99.03 | | 95.14 | |
| | CI | 98.06, 98.16 | | 90.32, 90.80 | | 98.92, 98.98 | | 90.34, 90.78 | | 98.93, 98.98 | | 94.62, 94.89 | |
| VI | μ | 97.86 | | 89.29 | | 98.81 | | 89.28 | | 98.82 | | 94.05 | |
| | σ | 0.04 | | 0.22 | | 0.02 | | 0.22 | | 0.02 | | 0.12 | |
| | ∧ | 97.78 | | 88.90 | | 98.77 | | 88.93 | | 98.77 | | 93.83 | |
| | ∨ | 97.92 | | 89.62 | | 98.85 | | 89.62 | | 98.85 | | 94.23 | |
| | CI | 97.83, 97.89 | | 89.16, 89.43 | | 98.80, 98.83 | | 89.14, 89.41 | | 98.80, 98.83 | | 93.98, 94.13 | |
| MCD | μ | 97.96 | | 89.78 | | 98.86 | | 89.82 | | 98.87 | | 94.32 | |
| | σ | 0.07 | | 0.37 | | 0.04 | | 0.34 | | 0.04 | | 0.20 | |
| | ∧ | 97.82 | | 89.10 | | 98.79 | | 89.20 | | 98.80 | | 93.94 | |
| | ∨ | 98.06 | | 90.31 | | 98.92 | | 90.30 | | 98.93 | | 94.62 | |
| | CI | 97.91, 98.00 | | 89.55, 90.00 | | 98.84, 98.89 | | 89.61, 90.03 | | 98.85, 98.90 | | 94.19, 94.45 | |
| | | \multicolumn{12}{c}{Undersampled ProBoost (method 1) with four levels} |
| | | FW | VW | FW | VW | FW | VW | FW | VW | FW | VW | FW | VW |
| VI | μ | 98.08 | 98.10 | 90.42 | 90.51 | 98.94 | 98.95 | 90.38 | 90.52 | 98.94 | 98.95 | 94.68 | 94.73 |
| | σ | 0.04 | 0.04 | 0.18 | 0.20 | 0.02 | 0.02 | 0.18 | 0.19 | 0.02 | 0.02 | 0.10 | 0.11 |
| | ∧ | 98.02 | 98.04 | 90.09 | 90.21 | 98.90 | 98.91 | 90.05 | 90.17 | 98.90 | 98.91 | 94.49 | 94.56 |
| | ∨ | 98.14 | 98.17 | 90.70 | 90.86 | 98.97 | 98.98 | 90.66 | 90.85 | 98.97 | 98.99 | 94.83 | 94.92 |
| | CI | 98.06, 98.11 | 98.08, 98.13 | 90.31, 90.53 | 90.38, 90.63 | 98.92, 98.95 | 98.93, 98.96 | 90.27, 90.49 | 90.40, 90.64 | 98.93, 98.95 | 98.93, 98.96 | 94.61, 94.74 | 94.66, 94.80 |
| | p | < 10⁻³ | < 10⁻³ | < 10⁻³ | < 10⁻³ | < 10⁻³ | < 10⁻³ | < 10⁻³ | < 10⁻³ | < 10⁻³ | < 10⁻³ | < 10⁻³ | < 10⁻³ |
| MCD | μ | 98.17 | 98.15 | 90.83 | 90.73 | 98.98 | 98.97 | 90.80 | 90.71 | 98.98 | 98.97 | 94.90 | 94.85 |
| | σ | 0.05 | 0.07 | 0.27 | 0.34 | 0.03 | 0.04 | 0.26 | 0.34 | 0.03 | 0.04 | 0.15 | 0.19 |
| | ∧ | 98.08 | 98.06 | 90.40 | 90.30 | 98.93 | 98.92 | 90.38 | 90.29 | 98.94 | 98.92 | 94.67 | 94.61 |
| | ∨ | 98.27 | 98.26 | 91.34 | 91.28 | 99.04 | 99.03 | 91.31 | 91.26 | 99.04 | 99.03 | 95.19 | 95.16 |
| | CI | 98.13, 98.20 | 98.10, 98.19 | 90.66, 90.99 | 90.52, 90.94 | 98.96, 99.00 | 98.95, 98.99 | 90.64, 90.96 | 90.50, 90.92 | 98.97, 99.00 | 98.95, 99.00 | 94.81, 94.99 | 94.73, 94.97 |
| | p | < 10⁻³ | < 10⁻³ | < 10⁻³ | < 10⁻³ | < 10⁻³ | < 10⁻³ | < 10⁻³ | < 10⁻³ | < 10⁻³ | < 10⁻³ | < 10⁻³ | < 10⁻³ |
| | | \multicolumn{12}{c}{Oversampled ProBoost (method 2) with four levels} |
| | | FW | VW | FW | VW | FW | VW | FW | VW | FW | VW | FW | VW |
| VI | μ | 98.23 | 98.26 | 91.17 | 91.31 | 99.02 | 99.03 | 91.11 | 91.26 | 99.02 | 99.04 | 95.09 | 95.17 |
| | σ | 0.02 | 0.05 | 0.12 | 0.25 | 0.01 | 0.03 | 0.13 | 0.25 | 0.01 | 0.03 | 0.06 | 0.14 |
| | ∧ | 98.18 | 98.18 | 90.92 | 90.89 | 98.99 | 98.99 | 90.85 | 90.85 | 98.99 | 98.99 | 94.96 | 94.94 |
| | ∨ | 98.26 | 98.34 | 91.32 | 91.68 | 99.04 | 99.08 | 91.27 | 91.64 | 99.04 | 99.08 | 95.18 | 95.38 |
| | CI | 98.22, 98.25 | 98.23, 98.29 | 91.09, 91.24 | 91.16, 91.47 | 99.01, 99.03 | 99.02, 99.05 | 91.03, 91.19 | 91.11, 91.42 | 99.01, 99.03 | 99.02, 99.05 | 95.05, 95.13 | 95.09, 95.26 |
| | p | < 10⁻³ | < 10⁻³ | < 10⁻³ | < 10⁻³ | < 10⁻³ | < 10⁻³ | < 10⁻³ | < 10⁻³ | < 10⁻³ | < 10⁻³ | < 10⁻³ | < 10⁻³ |
| MCD | μ | 98.12 | 98.13 | 90.60 | 90.66 | 98.96 | 98.96 | 90.78 | 90.81 | 98.97 | 98.97 | 94.78 | 94.81 |
| | σ | 0.03 | 0.05 | 0.14 | 0.26 | 0.02 | 0.03 | 0.13 | 0.19 | 0.01 | 0.03 | 0.08 | 0.14 |
| | ∧ | 98.06 | 98.05 | 90.28 | 90.27 | 98.92 | 98.92 | 90.55 | 90.59 | 98.94 | 98.94 | 94.60 | 94.59 |
| | ∨ | 98.16 | 98.21 | 90.80 | 91.05 | 98.98 | 99.01 | 90.99 | 91.08 | 98.99 | 99.01 | 94.89 | 95.03 |
| | CI | 98.10, 98.14 | 98.10, 98.16 | 90.52, 90.69 | 90.50, 90.82 | 98.95, 98.97 | 98.94, 98.98 | 90.70, 90.86 | 90.69, 90.92 | 98.96, 98.98 | 98.96, 98.99 | 94.73, 94.83 | 94.72, 94.90 |
| | p | < 10⁻³ | < 10⁻³ | < 10⁻³ | < 10⁻³ | < 10⁻³ | < 10⁻³ | < 10⁻³ | < 10⁻³ | < 10⁻³ | < 10⁻³ | < 10⁻³ | < 10⁻³ |

| | | Weighted ProBoost (method 3) with four levels | | | | | | | | | | | |
|---|---|---|---|---|---|---|---|---|---|---|---|---|---|
| | | FW | VW | FW | VW | FW | VW | FW | VW | FW | VW | FW | VW |
| VI | μ | 98.24 | 98.26 | 91.21 | 91.31 | 99.02 | 99.03 | 91.18 | 91.29 | 99.03 | 99.04 | 95.12 | 95.17 |
| | σ | 0.03 | 0.03 | 0.16 | 0.17 | 0.02 | 0.02 | 0.16 | 0.16 | 0.02 | 0.02 | 0.09 | 0.09 |
| | ∧ | 98.19 | 98.18 | 90.94 | 90.92 | 98.99 | 98.99 | 90.91 | 90.90 | 99.00 | 98.99 | 94.97 | 94.96 |
| | ∨ | 98.29 | 98.31 | 91.45 | 91.55 | 99.05 | 99.06 | 91.43 | 91.52 | 99.05 | 99.07 | 95.25 | 95.31 |
| | CI | 98.22, 98.26 | 98.24, 98.28 | 91.11, 91.31 | 91.20, 91.41 | 99.01, 99.03 | 99.02, 99.05 | 91.08, 91.28 | 91.19, 91.39 | 99.02, 99.04 | 99.03, 99.05 | 95.06, 95.17 | 95.11, 95.23 |
| | p | < 10$^{-3}$ | < 10$^{-3}$ | < 10$^{-3}$ | < 10$^{-3}$ | < 10$^{-3}$ | < 10$^{-3}$ | < 10$^{-3}$ | < 10$^{-3}$ | < 10$^{-3}$ | < 10$^{-3}$ | < 10$^{-3}$ | < 10$^{-3}$ |
| MCD | μ | 98.16 | 98.15 | 90.81 | 90.73 | 98.98 | 98.97 | 90.82 | 90.75 | 98.99 | 98.98 | 94.89 | 94.85 |
| | σ | 0.03 | 0.03 | 0.15 | 0.17 | 0.02 | 0.02 | 0.16 | 0.17 | 0.02 | 0.02 | 0.08 | 0.09 |
| | ∧ | 98.11 | 98.09 | 90.53 | 90.43 | 98.95 | 98.94 | 90.53 | 90.42 | 98.95 | 98.94 | 94.74 | 94.68 |
| | ∨ | 98.20 | 98.21 | 91.01 | 91.03 | 99.00 | 99.00 | 91.02 | 91.03 | 99.01 | 99.01 | 95.01 | 95.02 |
| | CI | 98.14, 98.18 | 98.13, 98.17 | 90.72, 90.90 | 90.63, 90.84 | 98.97, 98.99 | 98.96, 98.98 | 90.72, 90.92 | 90.64, 90.85 | 98.97, 99.00 | 98.96, 98.99 | 94.84, 94.94 | 94.79, 94.91 |
| | p | < 10$^{-3}$ | < 10$^{-3}$ | < 10$^{-3}$ | < 10$^{-3}$ | < 10$^{-3}$ | < 10$^{-3}$ | < 10$^{-3}$ | < 10$^{-3}$ | < 10$^{-3}$ | < 10$^{-3}$ | < 10$^{-3}$ | < 10$^{-3}$ |

TABLE 5
Performance of the VI and MCD models, when examining the first prepared dataset, using ProBoost with ten levels

| Model | Metric | Acc (%) | | | Sen (%) | | | Spe (%) | | | PPV (%) | | | NPV (%) | | | AUC (%) | | |
|---|---|---|---|---|---|---|---|---|---|---|---|---|---|---|---|---|---|---|---|
| | | Weighted ProBoost (method 3) with ten levels | | | | | | | | | | | | | | | | | |
| | | FW | VW | VWO | FW | VW | VWO | FW | VW | VWO | FW | VW | VWO | FW | VW | VWO | FW | VW | VWO |
| VI | μ | 97.67 | 97.68 | 97.74 | 88.34 | 88.42 | 88.72 | 98.70 | 98.71 | 98.75 | 88.34 | 88.39 | 88.71 | 98.71 | 98.72 | 98.75 | 93.52 | 93.57 | 93.73 |
| | ROI* | 5.43 | 6.08 | 8.52 | 5.43 | 6.08 | 8.52 | 5.43 | 6.08 | 8.51 | 5.61 | 6.02 | 8.63 | 5.39 | 6.15 | 8.49 | 5.43 | 6.08 | 8.52 |
| MCD | μ | 97.47 | 97.42 | 97.50 | 87.37 | 87.12 | 87.50 | 98.60 | 98.57 | 98.61 | 87.34 | 87.08 | 87.48 | 98.60 | 98.57 | 98.62 | 92.98 | 92.84 | 93.06 |
| | ROI* | 2.55 | 0.62 | 3.55 | 2.55 | 0.62 | 3.55 | 2.55 | 0.62 | 3.55 | 2.64 | 0.68 | 3.74 | 2.55 | 0.54 | 3.53 | 2.55 | 0.62 | 3.55 |

*ROI assessing the improvement from the FW model with four levels to a model with ten levels